\RequirePackage{fix-cm}
\documentclass[twocolumn]{svjour3}          
\smartqed  
\usepackage[utf8]{inputenc}
\usepackage{url}
\usepackage{cite}
\usepackage[colorlinks=true]{hyperref} 
\usepackage{amsmath,epsfig}
\usepackage{amssymb,amsfonts}
\usepackage{algorithmic}
\usepackage{graphicx}
\usepackage{textcomp}
\usepackage{xcolor}
\usepackage{multirow}
\usepackage{blindtext}
\usepackage{comment}
\usepackage{multicol}
\usepackage{multirow}
\usepackage{pdfpages}
\usepackage{subfigure}
\usepackage{booktabs}
\usepackage{tabularx}
\usepackage{adjustbox}
\usepackage{glossaries}
\graphicspath{ {./Figures/} }
\usepackage{array,multirow,makecell}
\usepackage{todonotes}
\usepackage{tikz}
\usetikzlibrary{positioning}
\usepackage{footmisc}
\usepackage{pgfplots, soul}
\pgfplotsset{width=10cm,compat=1.9}
\setcellgapes{1pt}
\makegapedcells
\newcolumntype{R}[1]{>{\raggedleft\arraybackslash }b{#1}}
\newcolumntype{L}[1]{>{\raggedright\arraybackslash }b{#1}}
\newcolumntype{C}[1]{>{\centering\arraybackslash }b{#1}}

\newcommand{\addcomment}[1]{{\color{black}{#1}}}

\usepackage{acronym}
\newacro{pgd}[PGD]{projected gradient descent}
\newacro{ae}[AE]{adversarial example}
\newacro{dnn}[DNN]{deep neural network}
\newacro{cnn}[CNN]{convolutional neural network}
\newacro{iid}[IID]{independent and identically distributed}
\newacro{fgsm}[FGSM]{fast gradient sign method}
\newacro{mim}[MIM]{momentum iterative method}
\newacro{nss}[NSS]{natural scene statistic}
\newacro{bm3d}[BM3D]{block matching 3D}
\newacro{eot}[EOT]{expectation over transformation}
\newacro{cw}[CW]{Carlini and Wagner}
\newacro{bpda}[BPDA]{backward pass differentiable approximation}
\newacro{lid}[LID]{local intrinsic dimensionality}
\newacro{fs}[FS]{feature squeezing}
\newacro{nic}[NIC]{neural-network invariant checking}
\newacro{mscn}[MSCN]{mean subtracted contrast normalized}
\newacro{ggd}[GGD]{generalized Gaussian distribution}
\newacro{aggd}[AGGD]{asymmetric generalized Gaussian distribution}
\newacro{svm}[SVM]{support vector machine}
\newacro{ssim}[SSIM]{Structural SIMilarity}

\newacro{relu}[ReLU]{rectified linear unit}
\newacro{roc}[ROC]{receiver operating characteristic}
\newacro{tpr}[TPR]{true positive rate}
\newacro{fpr}[FPR]{false positive rate}
\newacro{fp}[FP]{false positive}
\newacro{rrp}[RRP]{random resizing and padding}

\newacro{sfad}[SFAD]{Selective and Feature based Adversarial Detection}

\begin{document}

\title{Detect and Defense Against Adversarial Examples in Deep Learning using Natural Scene Statistics and Adaptive Denoising}

\titlerunning{Detect and Defense Against Adversarial Examples}

\author{Anouar Kherchouche         \and
        Sid Ahmed Fezza \and Wassim Hamidouche 
}


\institute{A. Kherchouche and W. Hamidouche \at
        Univ. Rennes, INSA Rennes, CNRS, IETR - UMR 6164, Rennes, France\\
         \email{kherchoucheanouar98@gmail.com, wassim.hamidouche@insa-rennes.fr}
				\and
				SA. Fezza \at
              National Institute of Telecommunications and ICT, Oran, Algeria \\
              \email{sfezza@inttic.dz}}

\date{Received: date / Accepted: date}

\maketitle

\begin{abstract}
Despite the enormous performance of \acp{dnn}, recent studies have shown their vulnerability to \acp{ae}, i.e., carefully perturbed inputs designed to fool the targeted \ac{dnn}. Currently, the literature is rich with many effective attacks to craft such \acp{ae}. Meanwhile, many defenses strategies have been developed to mitigate this vulnerability. However, these latter showed their effectiveness against specific attacks and does not generalize well to different attacks. 
In this paper, we propose a framework for defending \ac{dnn} classifier against adversarial samples. The proposed method is based on a two-stage framework involving a separate detector and a denoising block. The detector aims to detect \acp{ae} by characterizing them through the use of \ac{nss}, where we demonstrate that these statistical features are altered by the presence of adversarial perturbations. The denoiser is based on \ac{bm3d} filter fed by an optimum threshold value estimated by a \ac{cnn} to project back the samples detected as \acp{ae} into their data manifold. We conducted a complete evaluation on three standard datasets namely MNIST, CIFAR-10 and Tiny-ImageNet. The experimental results show that the proposed defense method outperforms the state-of-the-art defense techniques by improving the robustness\linebreak against a set of attacks under black-box, gray-box and white-box settings. The source code is available at: \url{https://github.com/kherchouche-anouar/2DAE}
\end{abstract}
\acresetall


\section{Introduction}
With the availability of large datasets \cite{deng2009imagenet,lin2014microsoft} in addition to the increase in computational power, the \acfp{dnn} have shown an outstanding performance in different tasks, such as image classification \cite{B0000,lecun1998gradient}, natural language processing \cite{B33333,cho2014learning}, object detection \cite{ren2015faster} and speech recognition \cite{hinton2012deep}. Despite their widespread use and  phenomenal success, 
these networks have shown that they are vulnerable to adversarial attacks. Szegedy {\it et al.} illustrated in \cite{szegedy2013intriguing} that small
and almost imperceptible perturbations added to a legitimate input image can easily fool the \acp{dnn} models and can make a misclassification with high confidence. The perturbed images are  called \acp{ae}. 

\begin{figure}[t!]
    \centering
\includegraphics[width=0.49\textwidth]{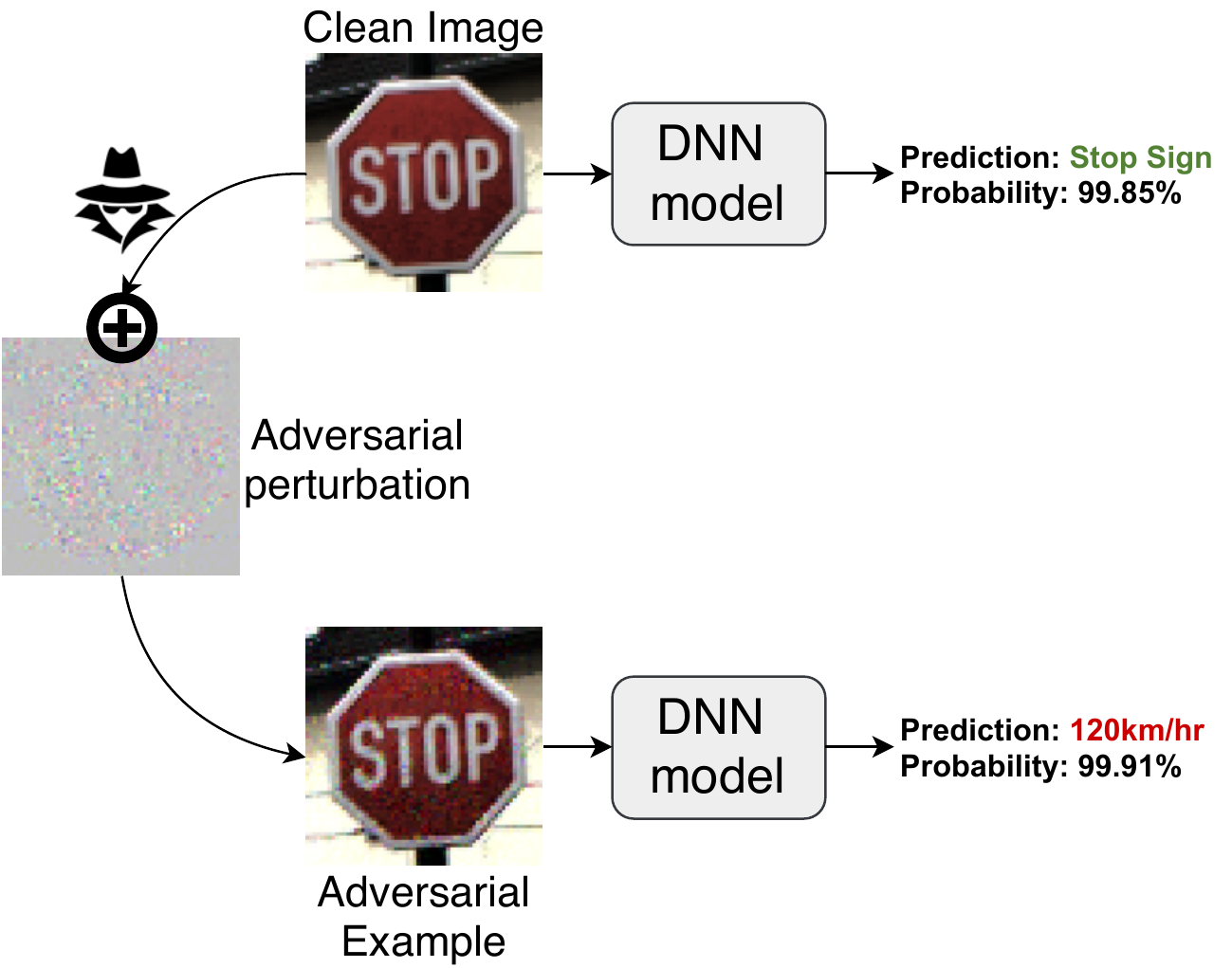}
    \caption{An example of adversarial attack in the context of street sign recognition. The introduction of a small imperceptible perturbation in the input image fools the \ac{dnn} classifier. The original image is classified as a \textit{Stop Sign} with 99.85\% confidence, while the adversarial example is classified as a \textit{120km/hr} with 99.91\% confidence.}
    \label{fig:figsys}
\end{figure}

These adversarial attacks raise an important issue of the robustness of \acp{dnn} against these attacks, which limits their use and can be an obstacle to deploy them in sensitive applications such as self-driving cars, healthcare, video surveillance, etc. For instance, in Figure \ref{fig:figsys}, the initially clean input image is classified correctly as a stop sign by the \ac{dnn} model with high confidence. When carefully crafted perturbations are added to the input image, this leads it misclassified as a 120km/hr, which is significantly dangerous and can cause fatal consequences. Therefore, it is of paramount importance to improve the robustness of \acp{dnn} models, especially, if they are deployed in such  critical applications.

\begin{figure}[t!]
    \centering
\includegraphics[width=0.45\textwidth]{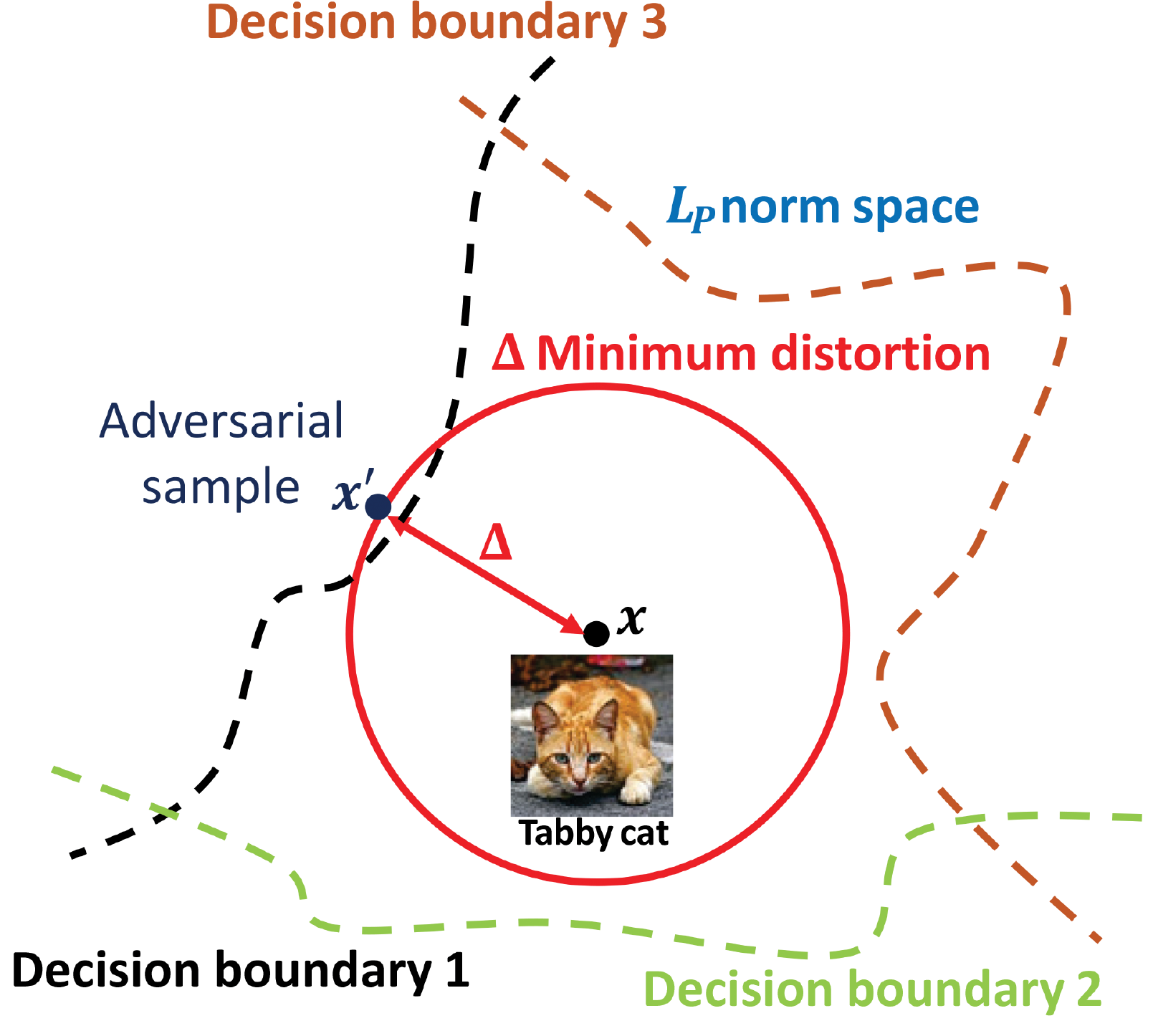}
    \caption{Data distribution over the manifold. We restrict the manifold of benign samples $x$ between decision boundaries (the true class), the attacked sample $x'$ is injected out from its data manifold by a minimum $L_p$ norm based distance $\Delta$.}
    \label{fig:fig1}
\end{figure}

Overall, unlike usual training, where the aim is to minimize the loss of \ac{dnn} classifier, an attacker tries to carefully craft an adversarial sample $x'$ by maximizing the loss with a small amount and thereby produce incorrect output. In other words, the attack tries to find the shortest adversarial direction $\Delta$ to inject the benign sample $x$ from its manifold into another one in order to mislead the \ac{dnn} model (see Figure \ref{fig:fig1}).

Various attacks are effective to generate an image-dependent \acp{ae} \cite{B16}, which limits its transferability to other images or models. Other methods \cite{moosavi2017universal} found some kind of universal perturbations that can fool a target classifier using any clean image. As shown in \cite{athalye2018synthesizing}, it is also possible
in the physical world to add 3D adversarial objects that can attack a \ac{dnn} model, thus creating a real security risk.

Consequently,  several defense methods have been proposed attempting  to  correctly  classify  \acp{ae} and\linebreak thereby  increasing  model’s  robustness. A defense\linebreak method aims to project the malicious sample $x'$ into it's data manifold to make the predicted label the same as the original sample $x$. Adversarial training \cite{madry2017towards,xie2019feature,carlini2018ground,kurakin2016adversarial,lee2017generative} is the most adopted technique that attempts to enhance the robustness against these vulnerabilities by integrating \acp{ae} to the clean ones into the training phase. However, such defense strategies do not generalize well against new/unknown attack models. Other defense methods try to reconstruct the \acp{ae} back to the training distribution by applying transformation on them. One of this approach consists to preprocess the \acp{ae} before feeding them to the \ac{dnn} model. For instance, denoising auto-encoders has been proposed in \cite{gu2014towards,bakhti2019ddsa} to remove/attenuate the perturbations. Similar approaches have been proposed based on generative models \cite{oord2016pixel}. These methods provided substantial results, however, systematically denoising each input image, can negatively impact the performance of clean images. Because, the denoiser can introduce blur if the denoising is incorrectly applied, which reduces the classification performance \cite{hendrycks2018benchmarking}. One of the possible solutions is to couple a defense strategy with a detection method as realized in the proposed work. Another limitation, generally these denoising-based approaches apply the denoising with a fixed non-adaptive strength, which is not optimal since certain adversarial perturbations are not distributed uniformly.

Almost all existing defense methods are effective against some specific attacks, but fail to defend new or more powerful ones, especially, when the attacker knows the details of defense mechanism \cite{B16,B24,B29}. Therefore, many recent works have focused on detecting \acp{ae} \cite{B24,B29,B25,B8,B9,B15,B22,B23,B26,B27,lu2017safetynet} instead. The detection of \acp{ae} may be useful to warn users or to take security measures in order to avoid tragedies. Furthermore, for online machine learning service providers, the detection can be exploited to identify malicious clients and reject their inputs \cite{B25}. However, as shown in \cite{B29}, the existing \acp{ae} detection methods reported high detection accuracy, but have also obtained high false positive rate, meaning that they reject a significant amount of clean images, which can be considered a failure of these detection approaches.

In this paper, we propose a novel approach for defending against \acp{ae}, which is classifier-agnostic, i.e., it is designed in such a way that can be used with any classifier without any changes. Our method is two-stage framework comprising a separate detector network and a denoiser block, where the sample detected as adversarial is fed to the denoiser network in order to denoise it. However, for the ones that are detected as clean, they are fed directly to the classifier model. The detector is based on \ac{nss}, where we rely on the assumption that the presence of adversarial perturbations alters some statistical properties of natural images. Thus, quantifying these statistical outliers, i.e., deviations from the regularity, using scene statistics enables the building of a binary classifier capable of classifying a given image as legitimate or adversarial. For the denoiser, we used the \ac{bm3d} filter in order to clean up the attacked image from the adversarial perturbation. \ac{bm3d} is one of the best denoiser algorithm allowing to tackle non-uniform adversarial perturbations. In addition, we built a \ac{cnn} model that predicts the adequate strength of denoising, i.e., the parameters filter values, which best mitigate \acp{ae}. The experimental results showed that the proposed detection method achieves high detection accuracy, while providing a low false positive rate. In addition, the obtained results showed that the proposed defense method outperforms the state-of-the-art defense techniques under the strongest black-box, gray-box and white-box attacks on three datasets namely MNIST, CIFAR-10 and Tiny-ImageNet.

The rest of this paper is organized as follows: \linebreak Section \ref{sec:sec2} reviews some attack techniques and defense methods that have been proposed in the literature. Section \ref{sec:proposal} describes the proposed approach. The experimental results are presented and analyzed in Section \ref{sec:experimental}. Finally, Section \ref{sec:conclusion} concludes the paper. Table \ref{tab:tab1} summarizes most of the notations used in this paper.

\begin{table}[t!]
\renewcommand{\arraystretch}{1.1} 
    \centering
    \caption{Notations used in the paper.}
		\adjustbox{max width=.49\textwidth}{%
\begin{tabular}{@{}ll@{}}
\toprule
Notations        & Description                                         \\ \midrule
$\xi$                         & Image space.                                        \\\hline
$I$                & Input image. \\\hline
$H \times W \times C$         & Height, width, channels of an image.                \\\hline
$x$                           & An instance of clean image.          \\\hline
$x^\prime$                          & A modified instance of $x$, adversarial image.    \\\hline
$\tilde{x}$                  &  A filtered instance of $x^\prime$, denoised image.                                     \\\hline
$c$                           & The true class label of an input instance $x$.   \\\hline
 $f_\theta(x)$                    & \begin{tabular}[c]{@{}l@{}} Output of a classifier model $f$ parameterized\\
                                                               by $\theta$, refers specifically to the predicted
                                                               \\likelihood for class $c$. 
                                \end{tabular} \\\hline
$d(x, x^\prime)$              & Distance metric  between $x$ and $x^\prime$.                                  \\\hline
$\parallel \cdot \parallel_p$ & $L_p$ norm.                                         \\\hline
$\nabla_x$                      & Derivative with respect to $x$ (gradient).                                           \\\hline
$\mathcal{L}$                 & Loss function.                                      \\\hline
$\mathcal{D}$                & Our detector block.                                       \\\hline
$\mathcal{S}$                & Our denoiser block.                                        \\\hline
$\widehat{I}$                & \begin{tabular}[c]{@{}l@{}} The \ac{mscn}\\  coefficients of image $I$.\end{tabular} \\\hline 
$\beta$                      & The shape parameter.                                    \\\hline
$\sigma^2$                    & The variance of a probability density function.                                 \\\hline
$\eta$                         & The Mean of the \acs{aggd}.                                              \\\hline
$\tau_{3 D}$                  & 3D  linear  transform.                              \\\hline
$\lambda_{3 D}$               & Threshold parameter of \acs{bm3d} filter.                               \\\hline
$f_X(x)$ & Probability density function. \\ \bottomrule
\end{tabular}
}
\label{tab:tab1}
\end{table}
\section{Related work}
\label{sec:sec2}
Adversarial examples are first introduced in this section, then different adversarial attacks are presented, and finally the state-of-the-art defense techniques are described. 
\subsection{Adversarial examples}
Given an image space $\xi = [0, 1]^{H \times W \times C}$, a target classification model $f_\theta(\cdot)$ and a legitimate input image  $x \in \xi$, an adversarial example is a perturbed image $x^\prime \in \xi$ such that $f(x^\prime) \neq f(x)$ and $d(x, x^\prime) \leq \epsilon$, where $\epsilon \geq 0$. $d$ is a distance metric used to measure the similarity between the perturbed and clean (unperturbed) input images~\cite{subQomex}. Three metrics are commonly used in the literature for generating \acp{ae} relying on $L_{p}$ norms, mainly $L_{0}$ distance, the Euclidean distance ($L_{2}$) and the {\it Chebyshev} distance ($L_{\infty}$ norm)~\cite{B16}.

\subsection{Adversarial Attacks}
Adversarial attacks fall into three main categories including black-box, gray-box and white-box attacks.\linebreak White-box attacks have a full access to both the defense technique and the target model's architecture and parameters, while black-box attacks have no access to the model's architecture and parameters. In this latter configuration, the attacker has only information on the output of the model (label or confidence score) for a given input. Finally, for gray-box  attacks also  referred as  semi  black-box  attacks, the  attacker is  unaware of the defense block, but has full access to the architecture and parameters of the model.

In the following, we describe three attacks considered in the evaluation of our defense method. These attacks are widely used in the literature to assess the performance of defense techniques. For more details on adversarial attacks, the reader can refer to the review paper on \acp{ae}~\cite{DBLP:journals/corr/abs-1911-05268}.

\subsubsection{\Acl{fgsm} attack}
Goodfellow {\it et al.}~\cite{B5} introduced a fast attack method called \acf{fgsm}. The \ac{fgsm} performs only one step gradient update along the direction of the sign of gradient at each pixel as follows
\begin{equation}
x^\prime = x + \epsilon \, sign (\nabla_x \mathcal{L}_\theta(x, y)),
\end{equation}
where $\theta$ is the set of model's parameters and $\nabla_x \mathcal{L}$ computes the first derivative (gradient) of the loss function $\mathcal{L}$ with respect to the input $x$. The $sign( \cdot )$ function returns the sign of its input and $\epsilon$ is a small scalar value that controls the perturbation magnitude. The authors proposed to bound the adversarial perturbation under the {\it Chebyshev} distance $|| x - x^\prime||_\infty < \epsilon$.

\subsubsection{\Acl{pgd} attack}
The \acf{pgd} attack was introduced by Madry {\it et al.} in~\cite{B18} to build a robust deep learning models with adversarial training. The authors formulated the generation of an \ac{ae} as a composition of an inner maximization problem and an outer minimization problem. 
 Specifically, they introduced the following saddle point optimization problem
\begin{equation}
\begin{split}
    & \min_\theta  \rho(\theta), \\    
    \text{where } \; & \rho(\theta) = \mathbf{E}_{(x,y)\sim D} [\max_{\delta \in S} \mathcal{L}_\theta(x+\delta,y)],
\end{split}
\end{equation}
with $\mathbf{E}$ is a risk function, $\delta$ is the magnitude of the perturbation and $S$ is a set of allowed perturbations.

The inner maximization is the same as attacking a neural network by finding an adversarial example that maximizes the loss. On the other hand, the outer minimization aims to minimize the adversarial loss.
\subsubsection{\acl{cw} attack}
Carlini and Wagner~\cite{B16} introduced an attack that can be used under three different distance metrics: $L_0$, $L_2$ and $L_\infty$. The \ac{cw} attack aims at minimizing a trade-off between the perturbation intensity $|| \delta ||_p$ and the objective function $g(x^\prime)$, with $x^\prime=x+\delta$ and $g(x^\prime) \leq 0$ if and only if $f(x^\prime)=c$ and $f(x)\neq c$
\begin{equation}
\begin{split}
     &\min_{\delta} || \delta ||_p + \lambda \, g(x+\delta), \\
     &\text{such that } \; x+\delta \in [0, 1]^n,
\end{split}
\label{CW}
\end{equation}
where $c$ is the target class and $\lambda>0$ is a constant calculated empirically through a binary search.

In the case of the $L_2$ norm, the problem in (\ref{CW}) can be expressed as follows
\begin{equation}
     \min_{\omega} \left |\left| \frac{1}{2} \left ( \tanh(\omega) +1 \right )-x \right | \right|_2^2 + \lambda \, g \left ( \frac{1}{2} \left ( \tanh(\omega) +1 \right ) \right ), \\
\label{CWL2}
\end{equation}
A change of variable introduces a new variable $\omega$ with $\delta = \frac{1}{2} \left ( \tanh(\omega) +1 \right ) - x$ that removes the constraint in~(\ref{CW}).

\subsection{Defenses against adversarial attacks}
\label{sec:DM}
As for the attacks, several defense techniques have been proposed in the literature in order to build more robust and resilient \acp{dnn} in \ac{ae}-prone context. Defending against adversarial attacks can fall into three categorizes: (1) adversarial training, (2) preprocessing, and (3) detecting \acp{ae}~\cite{B16b}.
\subsubsection{Adversarial training}
The adversarial training techniques consist in including \acp{ae} at the training stage of the model to build a robust classifier. Authors in \cite{DBLP:journals/corr/Moosavi-Dezfooli15,B4,liu2019rob} used benign samples with adversarial samples as data augmentation in the training process. In practice, different attacks can be used to generate the \acp{ae}. The optimized objective function can be formalized as a weighted sum of two classification loss functions as follows
\begin{equation}
    \lambda \; \mathcal{L}_\theta( x , c ) + (1- \lambda) \; \mathcal{L}_\theta( x^\prime,c ) \, 
\end{equation}
where $\lambda$ is a constant that controls the weighting of the loss terms between normal and \acp{ae}.  

In \cite{madry2017towards}, the authors showed that using only the \ac{pgd} attack for data augmentation can achieve state-of-the-art defense performance on both MNIST and CIFAR-10 datasets. However, as demonstrated in \cite{schmidt2018adversarially}, achieving a good generalization under adversarial training is hard to achieve, especially against an unknown attack.
\subsubsection{Preprocessing}
The defense techniques in the preprocessing category process the input sample before its classification by the model. Authors in \cite{DBLP:journals/corr/abs-1711-01991} proposed a defense based on \ac{rrp}. The objective of this method is to attenuate the adversarial perturbation and introduce randomness through the transformations applied on the input sample. This randomness at the inference makes the gradient of the loss with respect to the input harder to compute. It has been shown in~\cite{athalye2018obfuscated} that defense techniques relying on randomization \cite{liu2018towards,lecuyer2019certified,dwork2009differential,li2019certified,dhillon2018stochastic} are effective under black-box and gray-box attacks, but they fail against the worst case scenario of white-box attack. The total variance minimization and JPEG compression have been investigated in~\cite{DBLP:journals/corr/DziugaiteGR16} as preprocessing transformations to project back the \ac{ae} to its original data subspace. To break the effect of these transformations, Athalye {\it et al.} \cite{DBLP:journals/corr/AthalyeS17} proposed a method named \ac{eot} to create adversaries that fool defenses based on such transformations. Other defense techniques rely on denoising process to remove or alleviate the effect of adversarial perturbations. The first denoising-based defense technique was proposed in~\cite{meng2017magnet} as a stack of auto-encoders to mitigate the adversarial perturbations. However, it has been shown in~\cite{zantedeschi2017efficient} that this technique is vulnerable to transferable attack generated by \ac{cw} attack in black-box setting. Author in \cite{DBLP:journals/corr/abs-1710-10766} introduced PixelDefend defense, which aims to process an input sample before passing it to the classifier. The PixelDefend technique trains a generative model such as PixelCNN architecture \cite{oord2016pixel} only on clean data in order to approximate the distributions of the data. It has also been shown in \cite{DBLP:journals/corr/abs-1710-10766} that the generative model can be used to detect \acp{ae} by comparing the input sample to the clean data under the generative model.
Similar to PixelDefend, Defense-GAN method \cite{samangouei2018defense} trains a generator for an ultimate goal to learn the distributions of clean images. Therefore, at the inference, this method transforms the \acp{ae} by finding a close benign image based on its distribution. Authors in~\cite{buckman2018thermometer} proposed a thermometer encoding to break the linear extrapolation behavior of classifiers by processing the input with an extremely nonlinear function. ME-Net \cite{yang2019me} uses matrix-estimation techniques to reconstruct the image after randomly dropping pixels in the input image according to a \hbox{probability $p$.} \addcomment{Borkar \textit{et al.} \cite{borkar2020defending} introduced trainable feature regeneration units, which regenerate activations of vulnerable convolutional filters into resilient features. They have shown that regenerating only the top 50\% ranked adversarial susceptible features in a few layers is enough to restore their robustness.}

\begin{figure*}[t!]
    \centering
\includegraphics[scale=0.8]{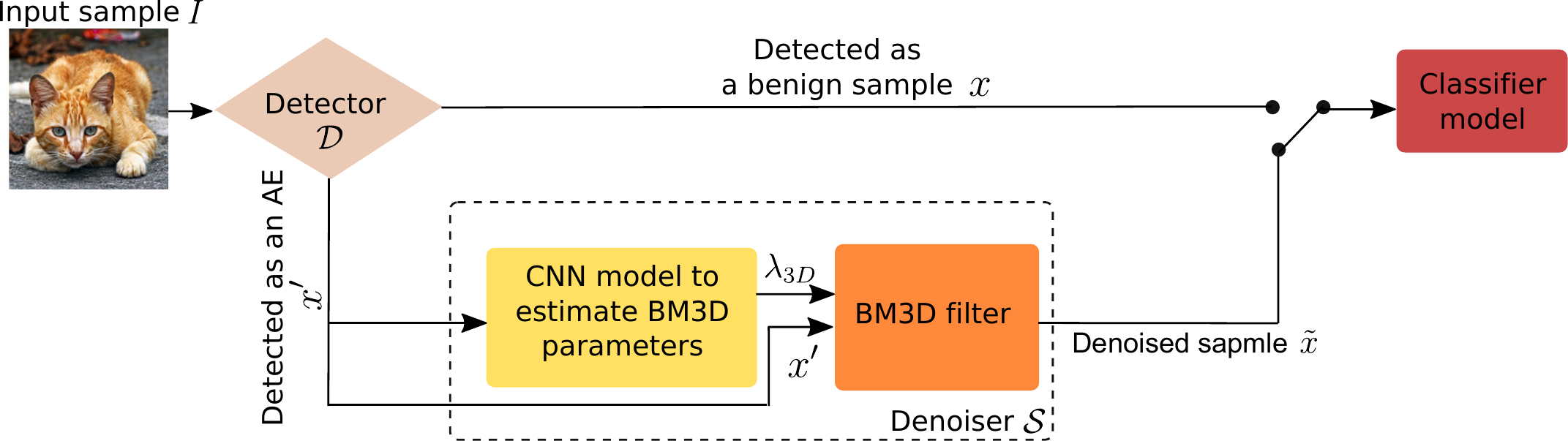}
    \caption{Overview of the proposed defense method workflow composed of detector $\mathcal{D}$ and denoiser $\mathcal{S}$ blocks.}
    \label{fig:fig2}
\end{figure*}

The gradient masking and obfuscated gradients have been explored to design defense techniques robust \linebreak against gradient based attacks. Meanwhile, \ac{bpda} \cite{athalye2018obfuscated} technique was proposed as a differentiable approximation for the defended model to obtain meaningful adversarial gradient estimates.  The \ac{bpda} techniques enables to derive a differentiable approximation for a non-differentiable preprocessing transformation that can be explored with any gradient based attack. It has been shown that \ac{bpda} approximation breaks the majority of these preprocessing based defense techniques. 

\subsubsection{Detecting adversarial samples}
Instead of trying to classify \acp{ae} correctly, which is difficult to achieve, many contributions have focused on only detecting these \acp{ae}. Grosse \textit{et al.} \cite{B23} proposed a technique that rely on statistical hypothesis on the input image to detect \acp{ae}, where the distribution of \acp{ae} statistically diverges from the data distribution. This hypothesis is explored to distinguish adversarial distributions from legitimate ones. Ma \textit{et al.} \cite{B27} explored the \ac{lid} concept for characterizing the dimensional properties of adversarial regions. The authors empirically showed that \ac{lid} of \acp{ae} is significantly higher than \ac{lid} of clean samples. Thus, they used the \ac{lid} of images as features to train a machine learning classifier to detect \acp{ae}. Xu \textit{et al.} \cite{B25} proposed a detection technique called \ac{fs}. Two \ac{fs} methods were considered to remove non-relevant features from the input: color bit-depth reduction and spatial smoothing, both with local and non-local smoothing.  \ac{dnn}-based predictions of clean and squeezed samples are compared to detect \acp{ae}. More specifically, the input sample is labeled as adversarial when the $l_1$ distance between the two \ac{dnn} predictions of squeezed and unsequeezed samples is higher than a threshold value.  Ma \textit{et al.} \cite{B15} proposed also another detection approach named \ac{nic}. This latter exploits two invariants in the \ac{dnn} classifier structure: the provenance channel and the activation value distribution channel. The \ac{nic} detector leverage these inveriants extracted from the \ac{dnn} classifier to perform runtime detection of adversarial samples. \addcomment{Under assumption that adversarial inputs leave activation fingerprints, i.e., the neuron activation values of clean and AEs are different, Eniser {\it et al.}~\cite{eniser2020raid}  proposed a binary classifier that takes as inputs the differences in neuron activation values between clean and \acp{ae} inputs to detect if the input is adversarial or not. In~\cite{sheikholeslamiprovably}, a method for jointly training a provably robust classifier and detector was proposed. The authors proposed a verification scheme for classifiers with detection under adversarial settings. They extend the Interval Bound Propagation (IBP) method to account for robust objective, which enables verification of the network for provable performance guarantees. Authors in~\cite{aldahdooh2021selective} proposed  a detection technique called \ac{sfad}. This latter uses the recent uncertainty method called SelectiveNet~\cite{selective2019} and integrates three detection modules. The first is selective detection module, which is a threshold-based detection derived from uncertainty of clean training data using SelectiveNet. The second is confidence detection module, which is threshold-based detection derived from softmax probabilities of clean training data from \ac{sfad}'s classifiers. \ac{sfad}'s classifiers analyze the representative data of last $N$-layers as a key point to present robust features of input data using autoencoding, up/down sampling, bottleneck, and noise blocks. The last module is ensemble prediction, which is mismatch based prediction between the detector and the baseline deep learning classifiers.}

The described detectors showed some limitations \cite{B29}, for instance they are effective against some specific attacks and lack generalization ability against different types of attacks. Also, they can achieve high accuracy but at the cost of increasing the \acl{fpr}, thus rejecting considerable legitimate inputs, which is not desired in real sensitive applications.

\section{Proposed Approach}
\label{sec:proposal}
In this paper, we propose a framework for defending against \acp{ae} \addcomment{in the digital domain, e.g., when attacking a computer vision system}. The proposed method consists of two main components that process the input sample before passing it to the classifier as shown in Figure~\ref{fig:fig2}. First, a detector block $\mathcal{D}$ distinguishes between clean sample $x$ and adversarial sample $x'$, and then a denoising block $\mathcal{S}$ alleviates perturbations in a sample detected as \ac{ae}. 
In other words, this denoising block $\mathcal{S}$ aims to project the adversarial sample back into the manifold of $x$. Finally, the classifier is fed by a denoised sample in order to predict the sample label. These two blocks will be explored in more detail in the next two sections.  \vspace{-3mm}
\subsection{Detector}
The detector $\mathcal{D}$ block relies on the concept of \ac{nss}. We assume that clean images possess certain regular statistical properties that are altered by adding adversarial perturbations. Thus, by characterizing these deviations from the regularity of natural statistics using \ac{nss}, it is possible to determine whether the input image $x$ is benign or malicious.

In order to extract scene statistics from input samples, we consider the efficient spatial \ac{nss} model~\cite{B0}, referred to as \ac{mscn} coefficients. The \ac{mscn} coefficients of a given input image $I$ are computed as follows 
\begin{align}
    \widehat{I}(i,j)=\frac{I(i,j)-\mu (i,j)}{\sigma (i,j)+c},
    \label{eq:0}
\end{align} 
where $i$ and $j$ are the pixel coordinates, and $c$ is a tiny constant added to avoid division-by-zero. The local mean $\mu$ and local standard deviation $\sigma$ are computed by (\ref{eq:mean}) and (\ref{eq:std}), respectively. 
\begin{equation}
    \mu(i,j)=\sum_{k=-3}^{3}\,\sum_{l=-3}^{3}\,w_{k,l} I_{k,l}(i,j).
    \label{eq:mean}
\end{equation}
\begin{equation}
    \sigma(i,j)=\sqrt{\sum_{k=-3}^{3}\,\sum_{l=-3}^{3}\,w_{k,l}(I_{k,l}(i,j)-\mu (i,j))^2}. 
    \label{eq:std}
\end{equation}
where $w=\{w_{k,l}|k=-3,...,3,l=-3,...,3\}$ is a 2D circularly-symmetric Gaussian weighting function.
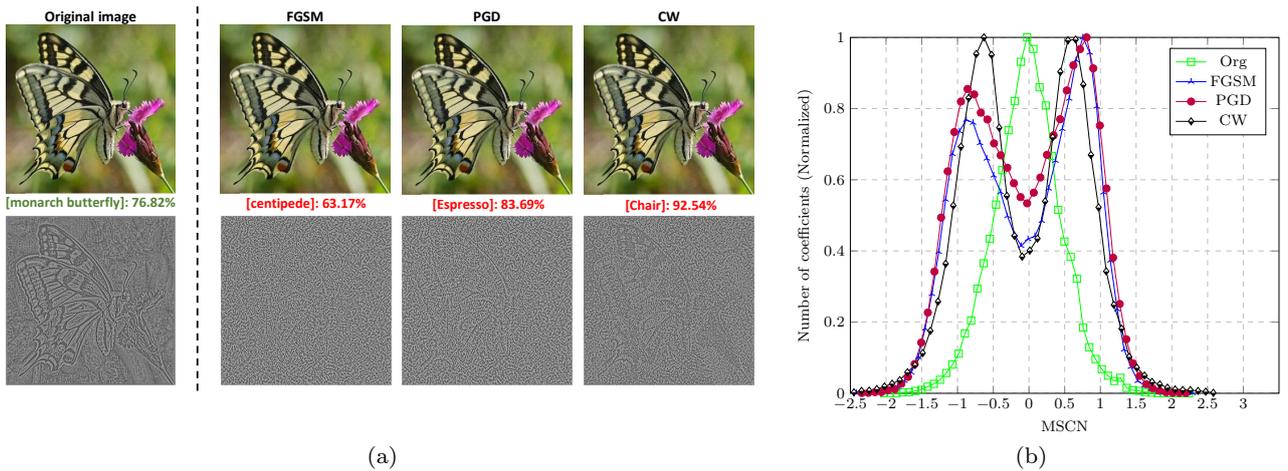
\begin{figure*}[t!]
\centering
\subfigure[]{\label{fig:fig3a}\includegraphics[scale=0.30]{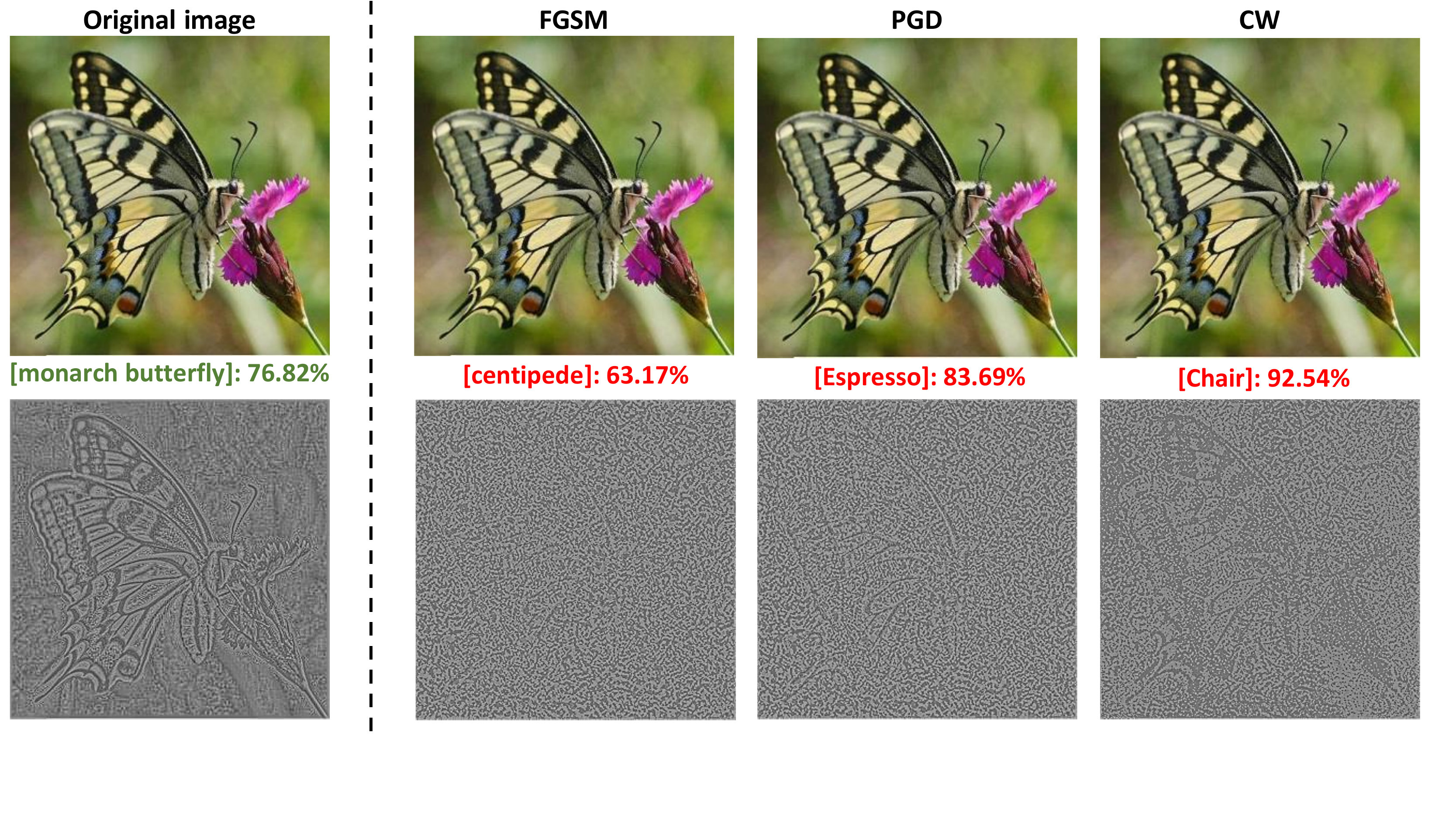}}\hspace{1.5mm}
\subfigure[]{\label{fig:fig3b}\begin{tikzpicture}[scale = 0.67]
\begin{axis}[
    title={},
    xlabel={MSCN},
    ylabel={Number of coefficients (Normalized)},
    xmin=-2.5, xmax=3.5,
    ymin=0, ymax=1,
    ytick={0,0.20,0.40,0.60,0.80,1.00},
    xtick={-2.5,-2,-1.5,-1,-0.5,0,0.5,1,1.5,2,2.5,3.0},
    legend pos=north east,
    ymajorgrids=true,
    xmajorgrids=true,
    grid style=dashed,
]
\addplot[
    color=green,
    mark=square,
    ]
    coordinates {(-2.0284075875919214, 0.0004048582995951417)
(-1.9413355866337088, 0.0006747638326585695)
(-1.854263585675496, 0.0006747638326585695)
(-1.7671915847172837, 0.002699055330634278)
(-1.680119583759071, 0.0032388663967611335)
(-1.5930475828008583, 0.006612685560053981)
(-1.505975581842646, 0.009716599190283401)
(-1.4189035808844332, 0.01862348178137652)
(-1.3318315799262206, 0.026585695006747637)
(-1.244759578968008, 0.03805668016194332)
(-1.1576875780097953, 0.056950067476383266)
(-1.0706155770515828, 0.08083670715249662)
(-0.9835435760933702, 0.11147098515519568)
(-0.8964715751351575, 0.16842105263157894)
(-0.8093995741769451, 0.20418353576248313)
(-0.7223275732187324, 0.2941970310391363)
(-0.6352555722605198, 0.36531713900134954)
(-0.5481835713023071, 0.4341430499325236)
(-0.46111157034409445, 0.5302294197031039)
(-0.374039569385882, 0.6269905533063428)
(-0.28696756842766935, 0.739136302294197)
(-0.1998955674694567, 0.8205128205128205)
(-0.11282356651124426, 0.9314439946018893)
(-0.025751565553031597, 1.0)
(0.06132043540518106, 0.9674763832658569)
(0.14839243636339372, 0.8596491228070176)
(0.23546443732160638, 0.8086369770580297)
(0.32253643827981904, 0.6670715249662618)
(0.40960843923803125, 0.5155195681511471)
(0.4966804401962439, 0.42645074224021595)
(0.5837524411544566, 0.3832658569500675)
(0.6708244421126692, 0.3219973009446694)
(0.7578964430708819, 0.18488529014844804)
(0.8449684440290945, 0.1292847503373819)
(0.9320404449873072, 0.096221322537112)
(1.0191124459455199, 0.06774628879892038)
(1.1061844469037325, 0.04939271255060729)
(1.1932564478619447, 0.03454790823211876)
(1.2803284488201574, 0.0437246963562753)
(1.36740044977837, 0.01524966261808367)
(1.4544724507365827, 0.008367071524966262)
(1.5415444516947954, 0.006747638326585695)
(1.628616452653008, 0.0035087719298245615)
(1.7156884536112207, 0.0017543859649122807)
(1.802760454569433, 0.0006747638326585695)
(1.8898324555276456, 0.0005398110661268556)
(1.9769044564858582, 0.0002699055330634278)
(2.0639764574440713, 0.0004048582995951417)
(2.1510484584022835, 0.0002699055330634278)
(2.2381204593604958, 0.0001349527665317139)};

\addplot[
    color=blue,
    mark=Mercedes star, 
    ]
    coordinates {(-2.414627164146805, 0.0003977724741447892)
(-2.3185700131204503, 0.0011933174224343676)
(-2.2225128620940957, 0.002386634844868735)
(-2.126455711067741, 0.0031821797931583136)
(-2.0303985600413865, 0.0053699284009546535)
(-1.9343414090150322, 0.009148766905330152)
(-1.8382842579886778, 0.017899761336515514)
(-1.7422271069623232, 0.035003977724741446)
(-1.6461699559359686, 0.05827366746221162)
(-1.550112804909614, 0.10182975338106603)
(-1.4540556538832594, 0.18178202068416865)
(-1.357998502856905, 0.2790373906125696)
(-1.2619413518305505, 0.39737470167064437)
(-1.1658842008041959, 0.544351630867144)
(-1.0698270497778415, 0.6722354813046937)
(-0.9737698987514869, 0.7360779634049324)
(-0.8777127477251323, 0.7684964200477327)
(-0.7816555966987777, 0.7603420843277645)
(-0.6855984456724231, 0.7028639618138425)
(-0.5895412946460687, 0.6595067621320605)
(-0.49348414361971416, 0.6119729514717581)
(-0.3974269925933598, 0.5654335719968179)
(-0.3013698415670052, 0.49781225139220364)
(-0.2053126905406506, 0.44749403341288785)
(-0.10925553951429601, 0.41408114558472553)
(-0.013198388487941415, 0.4335719968178202)
(0.08285876253841318, 0.44172633253778837)
(0.17891591356476777, 0.48448687350835323)
(0.2749730645911219, 0.5753778838504375)
(0.3710302156174765, 0.6533412887828163)
(0.4670873666438311, 0.7426412092283214)
(0.5631445176701857, 0.8277645186953063)
(0.6592016686965403, 0.9357597454256166)
(0.7552588197228949, 1.0)
(0.8513159707492495, 0.9570405727923628)
(0.9473731217756041, 0.8056881463802705)
(1.0434302728019587, 0.5644391408114559)
(1.1394874238283128, 0.37291169451073986)
(1.2355445748546674, 0.23607796340493237)
(1.331601725881022, 0.12291169451073986)
(1.4276588769073766, 0.07438345266507558)
(1.5237160279337312, 0.034606205250596656)
(1.6197731789600853, 0.022275258552108195)
(1.7158303299864404, 0.011933174224343675)
(1.8118874810127945, 0.004375497215592681)
(1.9079446320391495, 0.0027844073190135244)
(2.0040017830655037, 0.0015910898965791568)
(2.1000589340918587, 0.0005966587112171838)
(2.196116085118213, 0.0001988862370723946)
(2.292173236144567, 0.0005966587112171838)};
\addplot[
    color=purple,
    mark=otimes*,
    ]
    coordinates {(-2.341043936078205, 0.0009300162752848175)
(-2.248445867448091, 0.0011625203441060219)
(-2.1558477988179763, 0.0032550569634968614)
(-2.063249730187862, 0.0034875610323180655)
(-1.9706516615577476, 0.00906765868402697)
(-1.8780535929276332, 0.011857707509881422)
(-1.785455524297519, 0.026970471983259706)
(-1.6928574556674045, 0.04603580562659847)
(-1.60025938703729, 0.08230644036270635)
(-1.5076613184071757, 0.14252499418739828)
(-1.4150632497770612, 0.22692397116949548)
(-1.322465181146947, 0.3420134852359916)
(-1.2298671125168326, 0.4938386421762381)
(-1.1372690438867181, 0.6235759125784701)
(-1.044670975256604, 0.7340153452685422)
(-0.9520729066264895, 0.8198093466635666)
(-0.859474837996375, 0.8551499651243897)
(-0.7668767693662606, 0.8398046965821901)
(-0.6742787007361462, 0.788421297372704)
(-0.5816806321060319, 0.7693559637293652)
(-0.4890825634759175, 0.7021622878400372)
(-0.39648449484580306, 0.6686817019297838)
(-0.30388642621568884, 0.6338060916066031)
(-0.21128835758557418, 0.5900953266682167)
(-0.11869028895545997, 0.5512671471750755)
(-0.02609222032534575, 0.5338293420134852)
(0.06650584830476891, 0.5638223668914206)
(0.15910391693488313, 0.606370611485701)
(0.25170198556499734, 0.670076726342711)
(0.344300054195112, 0.7249476865845152)
(0.4368981228252262, 0.7700534759358288)
(0.5294961914553409, 0.850964891885608)
(0.6220942600854551, 0.9216461288072542)
(0.7146923287155693, 0.9667519181585678)
(0.807290397345684, 1.0)
(0.8998884659757982, 0.9130434782608695)
(0.9924865346059129, 0.7523831667054174)
(1.085084603236027, 0.575680074401302)
(1.1776826718661413, 0.38107416879795397)
(1.270280740496256, 0.25110439432690074)
(1.3628788091263702, 0.15182515694024645)
(1.4554768777564844, 0.08416647291327599)
(1.548074946386599, 0.04882585445245292)
(1.6406730150167133, 0.025807951639153684)
(1.7332710836468275, 0.013717740060451058)
(1.8258691522769421, 0.007207626133457335)
(1.9184672209070568, 0.0027900488258544524)
(2.0110652895371706, 0.0013950244129272262)
(2.1036633581672852, 0.0011625203441060219)
(2.1962614267974, 0.0009300162752848175)
};

\addplot[
    color=black,
    mark= halfdiamond*,
    ]
    coordinates {(-2.664373924929378, 0.0014678899082568807)
(-2.5573391470001186, 0.0033027522935779817)
(-2.450304369070859, 0.0027522935779816515)
(-2.3432695911415995, 0.006972477064220183)
(-2.23623481321234, 0.007706422018348624)
(-2.1292000352830804, 0.011926605504587157)
(-2.022165257353821, 0.01926605504587156)
(-1.9151304794245618, 0.027155963302752294)
(-1.8080957014953023, 0.03688073394495413)
(-1.7010609235660428, 0.05908256880733945)
(-1.5940261456367835, 0.07944954128440367)
(-1.486991367707524, 0.11302752293577982)
(-1.3799565897782644, 0.17559633027522936)
(-1.272921811849005, 0.25761467889908257)
(-1.1658870339197456, 0.36403669724770643)
(-1.0588522559904863, 0.5275229357798165)
(-0.9518174780612267, 0.6928440366972477)
(-0.8447827001319672, 0.8297247706422018)
(-0.7377479222027079, 0.953394495412844)
(-0.6307131442734484, 1.0)
(-0.5236783663441891, 0.9510091743119266)
(-0.4166435884149293, 0.7462385321100917)
(-0.30960881048567, 0.5548623853211009)
(-0.2025740325564107, 0.44238532110091744)
(-0.09553925462715096, 0.38403669724770645)
(0.011495523302108346, 0.40110091743119264)
(0.11853030123136765, 0.43486238532110094)
(0.22556507916062696, 0.5394495412844037)
(0.3325998570898867, 0.7192660550458716)
(0.439634635019146, 0.9031192660550459)
(0.5466694129484053, 0.9917431192660551)
(0.6537041908776651, 0.9939449541284404)
(0.7607389688069244, 0.8673394495412844)
(0.8677737467361837, 0.6697247706422018)
(0.9748085246654434, 0.5223853211009174)
(1.0818433025947027, 0.34275229357798165)
(1.188878080523962, 0.2486238532110092)
(1.2959128584532218, 0.181651376146789)
(1.402947636382481, 0.10256880733944954)
(1.5099824143117404, 0.07192660550458715)
(1.6170171922409997, 0.04990825688073394)
(1.724051970170259, 0.031926605504587154)
(1.8310867480995192, 0.025137614678899082)
(1.9381215260287785, 0.015963302752293577)
(2.045156303958038, 0.011009174311926606)
(2.152191081887297, 0.009174311926605505)
(2.2592258598165564, 0.009174311926605505)
(2.3662606377458157, 0.006788990825688073)
(2.473295415675076, 0.0045871559633027525)
(2.5803301936043352, 0.0014678899082568807)
};

\legend{Org,FGSM,PGD,CW}
    
\end{axis}
\end{tikzpicture}}
\caption{Illustration of the relationship between natural scene statistics and adversarial perturbations. (a) \textit{top}: the original image and different attacked versions.  \textit{bottom}: the \ac{mscn} coefficients of the images shown in the top row. (b) Histogram of \ac{mscn} coefficients for the original image and attacked images.}\label{fig:fig3}
\end{figure*}
\begin{figure*}[t!]
    \centering
\includegraphics[width=0.80\textwidth]{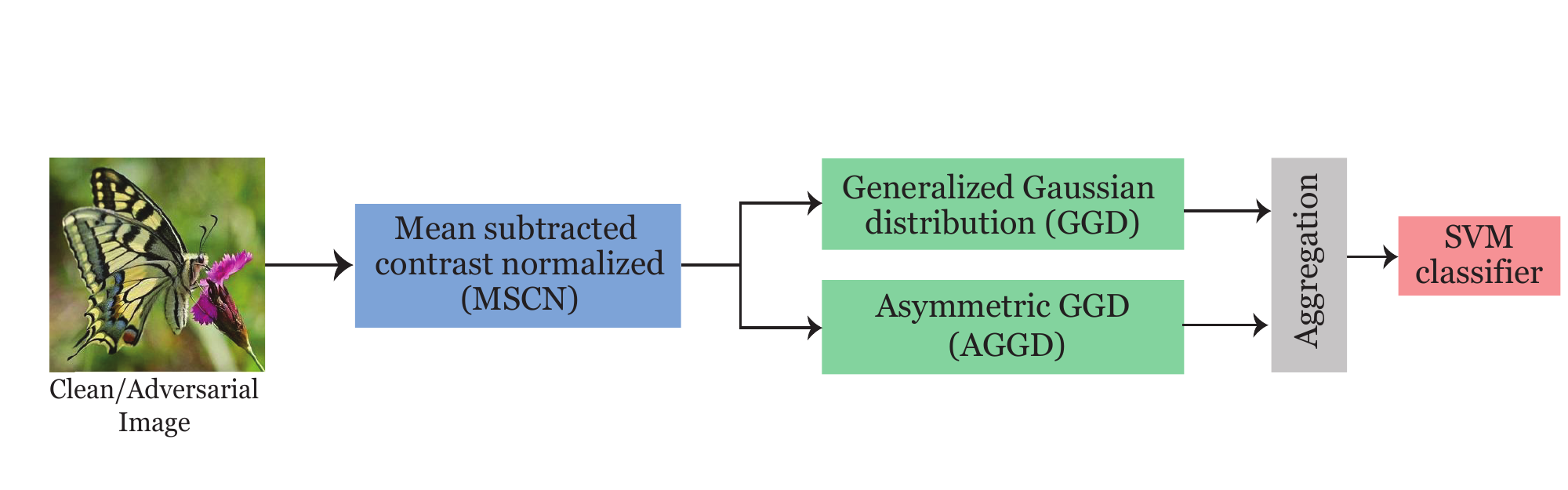}
    \caption{Overview of the proposed detection method workflow.}
    \label{fig:fig4}
\end{figure*}

In order to demonstrate that \ac{mscn} coefficients are affected by adversarial perturbations, Figure \ref{fig:fig3} illustrates the \ac{mscn} coefficients of the original (clean) image and its associated attacked versions. We consider three white-box attacks used in the experiments, namely \ac{fgsm}, \ac{pgd} and \ac{cw}. First, according to the obtained class label, it is clear that all attacks have succeeded in fooling the \ac{dnn} model with high confidence, while the attacked images are visually very close to the original one. From Figure~\ref{fig:fig3a}, we can also see that the \ac{mscn} coefficients of the original image differ significantly from those of adversarial attacks.     

In addition, in order to show how the \ac{mscn} coefficients vary with the presence of \acp{ae}, Figure \ref{fig:fig3b} plots the histogram of \ac{mscn} coefficients of images shown in Figure \ref{fig:fig3a} (top row). The original image exhibits a Gaussian-like \ac{mscn} distribution, while the same does not hold for the \acp{ae} which produce distributions with notable differences.

These results show that each attacked sample is characterized by its own histogram, which does not follow a Gaussian-like \ac{mscn} distributions like for the clean image. Based on that, we model these coefficients using the \ac{ggd} to estimate the parameters that are extracted from the scene statistics. The \ac{ggd} function is defined as follows
\begin{equation}
    f_X(x; \beta, \sigma^2) = \frac{\beta}{2\alpha\Gamma(1/\beta)}e^{-\big(\frac{|x|}{\alpha}\big)^\beta},
    \label{eq:1}
\end{equation} 
where
    $\alpha = \sigma \sqrt{\frac{\Gamma\big(\frac{1}{\beta}\big)}{\Gamma\big(\frac{3}{\beta}\big)}}$ 
and $\Gamma\left(\cdot\right)$ is the gamma function: $\Gamma \left( a \right) = \int\limits_0^\infty {t^{a - 1} e^{ - t} dt} \:, a >0$. 

The value of $\beta$ controls the shape and $\sigma^2$ is the variance controller parameter. Due to the symmetry property of the \ac{mscn} coefficients, we used the moment-matching~\cite{B2} to estimate the couple $(\beta, \, \sigma^2)$. To perform more accurate detection, we add the adjacent coefficients to model the pairwise products of neighboring \ac{mscn} coefficients along four directions (1) horizontal  $H$, (2) vertical  $V$, (3) main-diagonal  $D1$  and (4) secondary-diagonal $D2$~\cite{B0}. These orientations computed in Equation~(\ref{eq:orien}) have also certain regularities, which get altered in presence of adversary perturbations.
\begin{equation}
\begin{split}
    &H(i,j) = \hat{I}(i,j) \hat{I}(i, j + 1),
    \\&V(i,j) = \hat{I}(i,j) \hat{I}(i + 1, j),
    \\&D1(i,j) = \hat{I}(i,j) \hat{I}(i + 1, j + 1),
    \\&D2(i,j) = \hat{I}(i,j) \hat{I}(i + 1, j - 1).
\end{split}
\label{eq:orien}
\end{equation}
It is clear that the results of these pairwise products lead to an asymmetric distribution, so instead of using \ac{ggd}, we chose the \ac{aggd}, which is defined as follows
\begin{align}
    f_X(x; \nu, \sigma_l^2, \sigma_r^2) =  
   \begin{cases} 
      \frac{\nu}{(\alpha_l + \alpha_r)\Gamma\big(\frac{1}{\nu}\big)}e^{\big(-\big(\frac{-x}{\alpha_l}\big)^\nu\big)} & x < 0 \\
        \frac{\nu}{(\alpha_l + \alpha_r)\Gamma\big(\frac{1}{\nu}\big)}e^{\big(-\big(\frac{x}{\alpha_r}\big)^\nu\big)} & x \geq 0
\end{cases}
     \label{eq:7}
\end{align}
where
  $  \alpha_{side} = \sigma_{side} \sqrt{\frac{\Gamma\big(\frac{1}{\nu}\big)}{\Gamma\big(\frac{3}{\nu}\big)}}$
where $side$ can be either $r$ or $l$, $\nu$ represents the shape parameter and $\sigma_{side}^2$ expresses the left or the right variance parameters. So to estimate $(\nu ,{\sigma_l}^2,{\sigma_r}^2)$, we use the moment-matching as described in~\cite{B3}. Another parameter that is not reflected in the previous formula is the mean which is defined as follows
\begin{align}
    \eta = (\alpha_r - \alpha_l) \frac{\Gamma\big(\frac{2}{\nu}\big)}{\Gamma\big(\frac{1}{\nu}\big)}
     \label{eq:9}
\end{align}

The \ac{aggd} can be characterized by 4 features \- $(\eta,\nu,\sigma_l^2,\sigma_r^2)$ for each of the four pairwise products. The concatenation of the two \ac{ggd} parameters with the 16 \ac{aggd} ones results in 18 features ${\bf f}$ per image
\begin{multline*}
    {\bf f} = \{ \beta, \, \sigma^2, \, \eta_H, \, \nu_H, \, \sigma_{l_{H}}^2, \, \sigma_{r_{H}}^2, \, \eta_V, \, \nu_V, \, \sigma_{l_{V}}^2, \, \sigma_{r_{V}}^2, \, \\\eta_{D1}, \, \nu_{D1}, \, \sigma_{l_{D1}}^2, \, \sigma_{r_{D1}}^2, \, \eta_{D2}, \, \nu_{D2}, \, \sigma_{l_{D2}}^2, \, \sigma_{r_{D2}}^2 \}
     \label{eq:10}
\end{multline*}     

This low number of considered features motivates the choice to train a \ac{svm} binary classification model for the detector as shown in Figure \ref{fig:fig4}. Thus, the \ac{svm}  classifies each input image as either benign or \ac{ae}, where the sample detected as \ac{ae} is first processed by the denoiser, as detailed in the next section.

\subsection{Denoiser}
The aim of the denoiser block $\mathcal{S}$ is to alleviate the adversarial perturbations and thus project back the \ac{ae} into its original data manifold. In other words, the denoiser $\mathcal{S}$ tries to reconstruct from an adversarial example $x'$ a new sample $\tilde{x}$ such that $f_\theta(\tilde{x}) = f_\theta(x)$. 

The denoiser block processes only input samples detected as \ac{ae}, whereas, the samples detected as clean are directly passed to the classifier. In this way, we enhance the robustness against these adversarial attacks without affecting the classification accuracy of clean samples.
\begin{figure*}[t!]
    \centering
    \subfigure[MNIST]{\label{fig:fig5a}\includegraphics[scale=0.85]{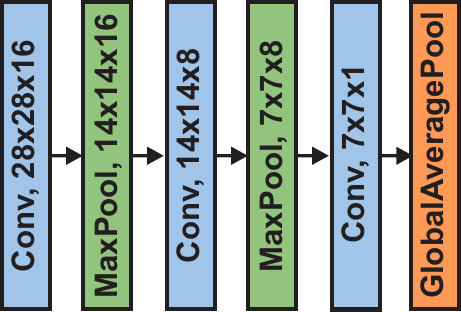}}\hspace{8mm}
    \subfigure[CIFAR-10]{\label{fig:fig5b}\includegraphics[scale=0.85]{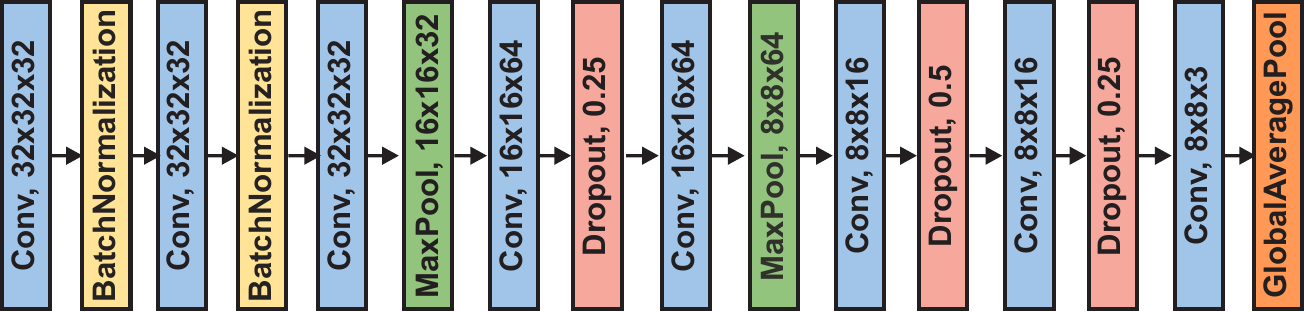}}
    \subfigure[Tiny-ImageNet]{\label{fig:fig5c}\includegraphics[scale=0.85]{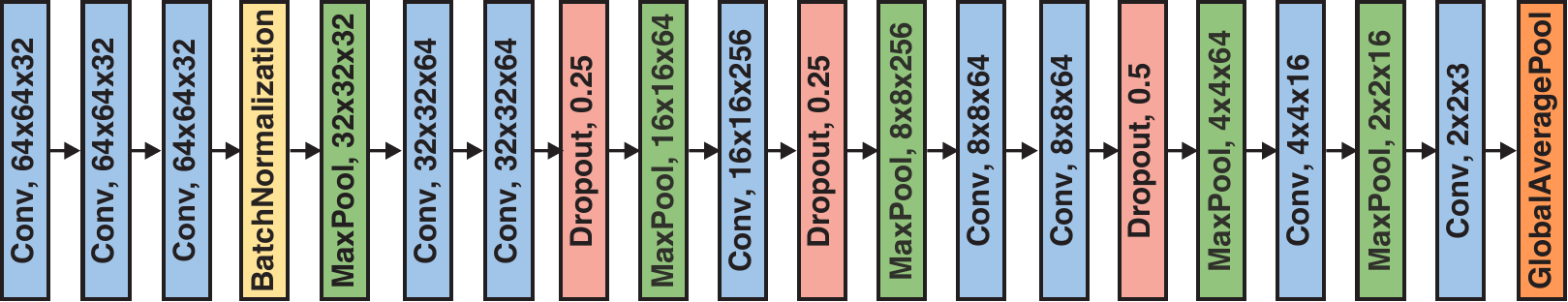}}
    \caption{The proposed \acs{cnn} architectures used in the prediction of  \acs{bm3d} parameters $\lambda_{3 D}$ for (a) MNIST, (b) CIFAR-10 and (c) Tiny-ImageNet datasets. The GlobalAveragePooling2D layer calculates the average of its input and outputs a single scalar value for each feature map.}
    \label{fig:fig5}
\end{figure*}

The denoiser relies on the \acf{bm3d} filter initially proposed in~\cite{dabov2007image}. This denoiser is considered to be one of the best non-learning-based denoising methods, furthermore, some work has shown that BM3D even out-performs deep learning-based denoising approaches for some real-world applications \cite{plotz2017benchmarking}. In addition, the BM3D allows locally adaptive parameter tuning based on block or region, making it suitable for non-uniform adversarial perturbations distribution.

The \ac{bm3d} filter first gathers similar 2D patches $P$ of an image in a 3D-block denoted $\mathbf{P}(P)$. For a given patch $P$ of size $\kappa \times \kappa$, the filter searches for similar patches $Q$ within a window of size $n \times n$ in the image, where $n>\kappa$. The search window is extracted such that the patch $P$ is the window center. The similarity between two patches is measured as follows 
\begin{align}
    \mathbf{P}(P)=\left\{Q: d(P, Q) \leq \tau \right\},
\end{align}
where $d$ is the normalized quadratic distance and $\tau$ is a threshold value set to check whether two patches are similar or not. In order to speed up the process, from the similar $Q$ patches within the 3D-block $\mathbf{P}(P)$, only the $N$ closest patches to $P$ are selected to get the $\tilde{\mathbf{P}}(P)$ 3D group, where $P$ is included. 

After the grouping step, a 3D linear transform $\tau_{3D}$ is applied on each 3D pile of correlated patches, followed by a shrinkage. Finally, the inverse of this isometric transform is applied to give an estimation of each patch as follows
\begin{align}
    \tilde {\mathbf{P}}(P)=\tau_{3 D}^{-1}\left[ \gamma\left(\tau_{3 D}\left[ \tilde{\mathbf{P}}(P) \right ] \right) \right],
\end{align}
where $\gamma$ is a thresholding that depends on $\lambda_{3 D}$:

\begin{align}
    \gamma(x)=\left\{\begin{array}{ll}
0 & \text { if } \quad|x| \leq \lambda_{3 D}  \\
x & \text { otherwise }
\end{array}\right.
\end{align}

The above grouping and filtering procedures are improved in a second step using Wiener filtering. This step is nearly the same as the first one, with only two differences. The first difference consists in comparing the filtered patches instead of the original ones at the grouping step. The second difference relies in using the Wiener filtering to process the new 3D groups, instead of using linear transform and thresholding. For further  details  about  the filtering process, the reader is referred to~\cite{dabov2007image}.

The performance of the \ac{bm3d} depends on its parameter settings. However, studies conducted in~\cite{bashar2016bm3d,lebrun2012analysis,mukherjee2019cnn} showed that the threshold $\lambda_{3 D}$ is the most crucial and significant parameter in \ac{bm3d}'s denoising process. Since \acp{ae} can contain different levels of adversarial perturbations, it is therefore important to choose the appropriate $\lambda_{3 D}$ parameter to mitigate each level of perturbation. To reach this goal, in this work, we use a \ac{cnn} to automatically predict the best $\lambda_{3 D}$ suited to each AE.

Inspired by the extension of \ac{bm3d} for color images initially investigated in~\cite{dabov2007image}, we propose to perform the grouping step relying only on the luminance component $Y$ after a color transformation from RGB color space to a luminance-chrominance color space, where $Y$ denotes luminance channel, while $U$ and $V$ refer to the two chrominance components. After building the 3D block on the $Y$ channel, we used it for all three channels, then the remaining \ac{bm3d}'s processes are  applied to each channel separately. Therefore, three parameters of the  thresholding $\lambda_{3 D}$ must be predicted, one per channel.

Thus, we performed the prediction of these thresholds under the RGB color space using a \ac{cnn} trained to derive, for each channel, an optimum value of $\lambda_{3 D}$. Figures~\ref{fig:fig5a}, \ref{fig:fig5b} and \ref{fig:fig5c} illustrate the three proposed architectures used to predict the optimal thresholding parameters for the three different image datasets, namely MNIST, CIFAR-10 and Tiny-ImageNet, respectively. These three different architectures have been proposed to adapt the \ac{cnn} to the complexity of parameters prediction which depends on the characteristics of the dataset including the dataset size, the color space and image resolution. The first architecture illustrated in Figure ~\ref{fig:fig5a} is fed with a grayscale image to predict a single $\lambda_{3 D}$ parameter, while architectures in Figures \ref{fig:fig5b} and \ref{fig:fig5c} are fed with the three RGB components of a color image to predict the three associated thresholding parameters, i.e., three $\lambda_{3 D}$ values.

The three networks are trained in a supervised learning fashion by minimizing a mean squared error loss function between the network output and the ground truth. The ground truth consists of a set of adversarial samples and the optimal denoising parameters enabling to back project an attacked image into its data manifold. To build the ground truth, an exhaustive denoising approach is conducted to denoise a set of adversarial samples perturbed by the \ac{pgd} attack at different $\epsilon$ magnitudes. The \ac{pgd} attack is selected based on the fact that adversarial training with \ac{pgd} attack tends to generalize well across a wide range of attacks \cite{B18}. So each sample is denoised with a set of  $\lambda_{3 D}$ parameters in the interval $[ 0.0, 1.0]$ with a step of $0.125$. This denoising process can result in several samples that are correctly classified. Among these samples, only one maximizing the \ac{ssim} \cite{wang2004image} image quality metric is retained in the ground truth. The \ac{ssim} metric is computed between the original image $x$ and the denoised one $\tilde{x}$.

The \ac{ssim} image quality metric assesses the quality of the denoised image with respect to the original one by exploring the structural similarity. Preserving the structural similarity after denoising will contribute to achieve a correct classification by the model with a high confidence score. The $\lambda_{3 D}$ parameters selected for each attacked sample with the highest \ac{ssim} score are assigned as the training labels for that adversarial sample.

\section{Experimental results}
\label{sec:experimental}
We describe in this section the evaluation process of our defense method with respect to the state-of-the-art defense techniques on three well known datasets: MNIST, CIFAR-10 and Tiny-ImageNet. First, we describe the selected datasets and the training stage, then the robustness of the proposed approach is assessed under three types of attacks, namely black-box, gray-box and white-box attacks.

\subsection{Datasets}
We evaluated the robustness of our defense technique on \acp{cnn} models trained on three standard datasets, namely:
\begin{itemize}
    \item \textbf{MNIST} dataset consists of grayscale hand-written 10 digits images of size $28\times28$. This dataset contains 70,000 images split into training, validation and testing sets with 50,000, 10,000 and 10,000 images, respectively. 
    \item \textbf{CIFAR-10} dataset contains color images of size $32\times32$. It has ten classes and 60,000 images divided into training and testing sets with 50,000 and 10,000 images, respectively. 
    \item \textbf{Tiny-ImageNet} dataset includes also color images of size $64\times64$ with a greater number of 200 classes. Each class includes 500, 50 and 50 images used for training, validation and testing, respectively.
\end{itemize}    

We built our own \ac{cnn} classifier for MNIST dataset resulting in a state-of-the-art accuracy of $99.4\%$. For CIFAR-10 and Tiny-ImageNet datasets, we considered existing models achieving accuracy scores of 98.5\%~\cite{cubuk2019autoaugment} and 69.2\%~\cite{tiny}, respectively.

\begin{table*}[t]
\renewcommand{\arraystretch}{1.1} 
\centering
\caption{Performance of the proposed defense method under black-box attacks.\label{tab:tab3}}
\begin{tabular}{l lcc cccc cc}
\toprule
\multicolumn{1}{l}{\multirow{2}{*}{\textbf{Dataset}}}&\multicolumn{1}{l}{\multirow{2}{*}{\textbf{Method}}} & \multicolumn{1}{c}{\multirow{2}{*}{\textbf{FGSM}}} & \multicolumn{4}{c}{\textbf{\acs{pgd}}}                   & \multicolumn{2}{c}{\textbf{CW}}   \\ 
\cmidrule(lr){4-7} 
\cmidrule(lr){8-9} 
\multicolumn{1}{c}{} &\multicolumn{1}{c}{}                                 & \multicolumn{1}{l}{}                              & \textbf{7 steps}     & \textbf{20 steps}& \textbf{40 steps}     & \textbf{100 steps}     & \textbf{$\kappa=20$}     & \textbf{$\kappa=50$}     \\ \midrule
\multicolumn{1}{l}{\multirow{4}{*}{\rotatebox[origin=c]{0}{MNIST}}}& Madry \cite{madry2017towards}                                               & 96.8\% &-&-                                            & 96.0\%          & 95.7\%               &96.4\%              & 97.0\%                       \\ 
&Thermometer \cite{buckman2018thermometer}                                         & -&-&-                                                  & 41.1\%          & -               & -              & -                              \\ 
& ME-Net \cite{yang2019me}                                               & 93.2\% &-&-                                            & 92.8\%          & 92.2\%          & 98.8\%                  & {98.7\%} \\ 
& {Our method}                                  & \textbf{97.6\%} &\textbf{99.4\%} & \textbf{99.4\%}                                   & \textbf{99.4\%} & \textbf{99.4\%} & \textbf{99.2\%} & \textbf{98.9\%}           \\ \bottomrule
\multicolumn{1}{l}{\multirow{4}{*}{\rotatebox[origin=c]{0}{CIFAR-10}}}&Madry \cite{madry2017towards}                                               & 67.0\%                                             & 64.2\%          & -               & -  &-             & 78.7\%          & -               \\ 
& Thermometer \cite{buckman2018thermometer}                                         & -                                                  & 77.7\%          & -               & - &-             & -               & -               \\ 
& ME-Net \cite{yang2019me}                                                & 92.2\%                                             & 91.8\%          & 91.8\%          & 91.3\%&-         & 93.6\%          & \textbf{93.6\%} \\ 
&{Our method}                                  & \textbf{95.7\%}                                    & \textbf{98.3\%} & \textbf{98.2\%} & \textbf{98.2\%} &\textbf{97.9\%}& \textbf{93.8\%} & 92.8\%          \\ \bottomrule
\multicolumn{1}{l}{\multirow{2}{*}{\rotatebox[origin=c]{0}{Tiny-ImageNet}}}&ME-Net \cite{yang2019me}                                               & 67.1\%                                             & 66.3\%          & 60.0\%          & 65.8\% &-        & 67.6\%          & \textbf{67.4\%} \\ 
& {Our method}                                  & \textbf{67.7\%}                                    & \textbf{68.8\%} & \textbf{69.1\%} & \textbf{69.1\%} &\textbf{68.9\%}& \textbf{68.3\%} & 67.2\%          \\ \bottomrule
\end{tabular}
\end{table*}

\subsection{Training process}
{ \bf Detector }
The detector is trained with a blend of clean and attacked samples. For MNIST dataset, we have selected 1,000 clean samples and generated adversarial samples with the \ac{pgd} attack. 
These 1,000 adversarial images with the associated 1,000 clean images are used for the training. The same process is carried-out for CIFAR-10 and Tiny-ImageNet datasets. The obtained features from each sample are provided as inputs to the \ac{svm} classifier. The Sigmoid kernel is used in the \ac{svm} model since it achieves a good accuracy for non-linear binary classification problems.  \newline 
{ \bf Denoiser } A separate \ac{cnn} model is trained to estimate the denoising parameters for each dataset. As described previously, the denoiser block deals only with attacked samples, based on that, we generated 10,000 samples with the \ac{pgd} attack.  The three \ac{cnn} architectures for MNIST, CIFAR-10 and Tiny-ImageNet datasets are shown in Figures \ref{fig:fig5a}, \ref{fig:fig5b} and \ref{fig:fig5c}, respectively. We used the \ac{relu} as an activation function after each convolution layer. Some dropout layers are added to networks of color image datasets, i.e, CIFAR-10 and Tiny-ImageNet, with different rate to prevent over-fitting.  
The \acp{cnn} of CIFAR-10 and Tiny-ImageNet datasets include batch normalization layers to stabilize and accelerate the learning process. We used for the three architectures a learning rate of 0.01 and a large momentum of 0.9. Finally, the architectures are trained using 64 epochs with a batch-size of 128 for MNIST and CIFAR-10 datasets, while 128 epochs with a batch-size of 32 for Tiny-ImageNet dataset. 
\subsection{Results and analysis}
The performance of the proposed defense method are evaluated under $l_{\infty}$ bounded attacks~\cite{madry2017towards,buckman2018thermometer,yang2019me,song2017pixeldefend}. We compare our method with three state-of-the-art defense techniques on the three considered datasets under black-box, gray-box and white-box attacks, as recommended in~\cite{carlini2019evaluating}. The chosen defense techniques include one of the best performing adversarial training defenses developed by  Madry {\it et al.}~\cite{madry2017towards} and two prepossessing methods including Thermometer~\cite{buckman2018thermometer} and ME-Net~\cite{yang2019me}. Furthermore, we investigate the effectiveness of the proposed detector block  to detect the \acp{ae}.

\subsubsection{Defense block performance}
The robustness of the proposed defense method is assessed against three attacks: \acf{fgsm}, \acf{pgd} and \acf{cw}. The \ac{cw} implementation~\cite{B16} provided by the authors is used, while the implementations of \ac{fgsm} and \ac{pgd} attacks are from the open-source {\it CleverHans} library~\cite{cleverhans}.  
\newline 
\textbf{Black-box attacks}
This kind of attack is performed to fool a classifier model when an attacker can not perform back propagation to generate adversarial samples from the network model. Based on this, we train substitute networks for MNIST, CIFAR-10 and Tiny-ImageNet datasets to generate \acp{ae} with \ac{fgsm}, \ac{pgd} and \ac{cw} attacks. We set the attacks hyper-parameters as in \cite{yang2019me}, where we use for MNIST a perturbation magnitude $\epsilon$ of  $0.3$ for both \ac{fgsm} and \ac{pgd}. This latter is used with two iteration configurations: 40 and 100 steps. Regarding CIFAR-10 and Tiny-ImageNet datasets, the \ac{pgd} attack is used with a magnitude  perturbation $\epsilon$ of $0.03$ and four iteration configurations: 7, 20, 40 and 100 steps. 
For the \ac{cw} attack, we consider two different confidence values of $\kappa=20$ and $\kappa=50$.
\begin{table*}[t]
\centering
\renewcommand{\arraystretch}{1.2} 
\caption{Performance of the proposed defense method under gray-box attacks.\label{tab:tab4}}
\begin{tabular}{l lcc cccc cc}
\toprule
\multicolumn{1}{l}{\multirow{2}{*}{\textbf{Dataset}}}&\multicolumn{1}{l}{\multirow{2}{*}{\textbf{Method}}} & \multicolumn{1}{c}{\multirow{2}{*}{\textbf{FGSM}}} & \multicolumn{4}{c}{\textbf{\acs{pgd}}}                   & \multicolumn{2}{c}{\textbf{CW}}   \\ 
\cmidrule(lr){4-7} 
\cmidrule(lr){8-9} 
\multicolumn{1}{c}{} &\multicolumn{1}{c}{}                                 & \multicolumn{1}{l}{}                              & \textbf{7 steps}     & \textbf{20 steps}& \textbf{40 steps}     & \textbf{100 steps}     & \textbf{$\kappa=20$}     & \textbf{$\kappa=50$}     \\ \midrule
\multicolumn{1}{l}{\multirow{2}{*}{\rotatebox[origin=c]{0}{MNIST}}}&ME-Net \cite{yang2019me}  &-&-                                               &96.2\% &95.9\%& 95.3\%&{98.8\%}&\textbf{98.7\%}\\ 
& {Our method}                                  & \textbf{97.4\%} &  \textbf{99.4\%} & \textbf{99.3\%}                                 & \textbf{99.3\%} & \textbf{99.3\%} & \textbf{99.1\%} & {98.5\%}           \\ \bottomrule
\multicolumn{1}{l}{\multirow{2}{*}{\rotatebox[origin=c]{0}{CIFAR-10}}}&ME-Net \cite{yang2019me}                                               &85.1\%                                             & 84.9\%          & 84.0\%         &82.9\%&- & \textbf{84.0}\%         & \textbf{77.1\%}          \\ 
& {Our method}                                  & \textbf{88.4\%}                                    & \textbf{95.0\%} &\textbf{94.8\%}& \textbf{94.8\%} &\textbf{94.5\%}& 83.7\% & 75.8\%  \\ \bottomrule
\multicolumn{1}{l}{\multirow{2}{*}{\rotatebox[origin=c]{0}{Tiny-ImageNet}}}&ME-Net \cite{yang2019me} 
&{66.5\%} &64.0\%& 62.6\%& 59.2\%&-&{58.3\%}&{58.2\%}\\ 
&{Our method} &\textbf{67.7\%}&\textbf{69.0\%}&\textbf{68.9\%}&\textbf{68.9\%}&\textbf{68.5\%}&\textbf{61.2\%} &\textbf{60.1\%}\\ \bottomrule
\end{tabular}
\end{table*}

Table \ref{tab:tab3} gives the performance of our defense method compared to the selected defenses under black-box attacks on the three datasets. We can notice that the proposed defense achieves the highest accuracy performance on the three datasets, except against \ac{cw} attack under $\kappa=50$ on the two color datasets. Moreover, most defense techniques are robust against the considered attacks on MNIST dataset, except the Thermometer defense which achieved the lowest accuracy of 41.1\%. We can also note, in MNIST dataset, that Madry's defense performs better than ME-Net which obtained $92.8\%$ and $92.2\%$ accuracy against \ac{pgd} attack at 40 and 100 steps, respectively. This can be explained by the fact that the \ac{pgd} \acp{ae} were included in the training set of the Madry defense. However, the preprocessing conducted in the ME-Net allows this method to achieve relatively higher classification accuracy for \ac{cw} attack compared to Madry's method. This is particularly true for CIFAR-10 dataset where ME-Net outperforms Madry's defense for all attacks considered. Finally, it is clear that our defense outperforms the considered defenses, for instance, it obtained the highest accuracies of $97.6\%$ and $98.9\%$ against FGSM and CW ($\kappa=50$) attacks, respectively. In addition, the proposed method achieves a constant accuracy of $99.4\%$ against \ac{pgd} attack despite the  change in the number of iterations.
\newline 
\textbf{Gray-box attacks}
In this setting, an adversary has knowledge of the hyper-parameters of the classifier without any information on  the defense technique. 
This kind of attacks are much stronger than black-box attacks in reducing the robustness of the defense mechanism. Table~\ref{tab:tab4} compares the accuracy performance of the proposed defense against gray-box attacks with respect to ME-Net technique on the three datasets. We can first of all notice that the  accuracy is lower compared to the black-box attacks. However, our defense method outperforms the ME-Net defense at all attacks, except against \ac{cw} attack where similar performance is reported. 
\newline 
\textbf{White-box attacks} 
In white-box attacks, an adversary has a full access to all hyper-parameters of the classifier architecture and the defense technique. We assessed the robustness of our defense method against such strong attacks using the \acf{bpda} attack~\cite{athalye2018obfuscated}. The proposed defense technique consists of a preprocessing method which is independent from the classifier model as shown in Figure~\ref{fig:fig2}. This causes gradient masking for gradient-based attacks. In other words, on the backward pass the gradient of the preprocessing step is non-differentiable and therefore useless for gradient-based attacks. The \ac{bpda} approximates the gradient of non-differentiable blocks, which makes it useful for evaluating  our defense technique against white-box attacks.
\begin{table*}[t!]
\centering
\renewcommand{\arraystretch}{1.2} 
\caption{Performance of the proposed defense method against \ac{bpda}-based \ac{pgd} under white-box attack at up to 1000 steps on the three datasets.\label{tab:tab5}}
\begin{tabular}{l l c ccccc}
\toprule
\multicolumn{1}{l}{\multirow{2}{*}{\textbf{Dataset}}} &\multicolumn{1}{c}{\multirow{2}{*}{\textbf{Method}}} &  \multicolumn{5}{c}{\textbf{Attack steps}}\\ 
\cmidrule(lr){3-7} 

\multicolumn{1}{c}{}&\multicolumn{1}{c}{} & \textbf{7}      & \textbf{20}     & \textbf{40}     & \textbf{100}     & \textbf{1000}     \\ \midrule
\multicolumn{1}{l}{\multirow{3}{*}{\rotatebox[origin=c]{0}{MNIST}}}& Madry \cite{madry2017towards}&-& -&  93.2\% & 91.8\% & 91.6\% \\ 
&ME-Net \cite{yang2019me} &-&-& 94.0\% & 91.8\% & 91.0\%\\ 
&{Our method}&-&-&\textbf{95.3\%}&\textbf{94.9\%} &\textbf{94.7\%}\\ \bottomrule

\multicolumn{1}{l}{\multirow{3}{*}{\rotatebox[origin=c]{0}{CIFAR-10}}}& Madry \cite{madry2017towards}&  50.0\% & 47.1\% & 47.0\% & 46.9\% & 46.8\% \\ 
&ME-Net \cite{yang2019me}  & \textbf{74.1\%} & 61.6\% & 57.4\% & 55.9\% & 55.1\%\\ 
&{Our method}&{71.2\%}&\textbf{70.6\%}&\textbf{70.3\%}&\textbf{69.9\%} &\textbf{69.6\%}\\ \bottomrule

\multicolumn{1}{l}{\multirow{3}{*}{\rotatebox[origin=c]{0}{Tiny-ImageNet}}}& Madry \cite{madry2017towards}& 23.3\% & 22.4\% & 22.4\% & 22.3\% & 22.1\% \\ 
&ME-Net \cite{yang2019me} & \textbf{38.8\%} & 30.6\% & 29.4\% & 29.0\% &28.5\%\\ 
& {Our method}&36.0\%&\textbf{35.8\%}&\textbf{35.7\%}&\textbf{35.2\%} &\textbf{34.8\%}\\ \bottomrule
\end{tabular}
\end{table*}

Thereby, we approximate the denoiser block with the identity function $g(x) = x$, which is often effective as reported in~\cite{tramer2020adaptive}. This approach enables to approximate the true gradient and thus to bypass the defense, which allows using a standard gradient-based attack based on \ac{bpda}. Similar to \cite{yang2019me}, we selected the \ac{bpda}-based \ac{pgd} attack to assess the efficiency of our method. Table~\ref{tab:tab5} gives the accuracy performance on the three datasets for different steps of the \ac{bpda}-based \ac{pgd} attack. We can notice that our defense outperforms Madry and ME-Net defense techniques on MNIST dataset. On CIFAR-10 and Tiny-ImageNet\linebreak datasets, the accuracy scores of our defense technique are higher than those of  Madry and ME-Net defenses, except in the 7 iterations configuration. At this configuration, ME-Net defense outperforms our solution by around 3\% benefiting mainly from its randomness operation. However, the ME-Net defense performance would be lower against the \ac{eot} technique that estimates the gradient of random components~\cite{athalye2018obfuscated}. 
\begin{figure}[t!]
    \centering
\includegraphics[scale=0.6]{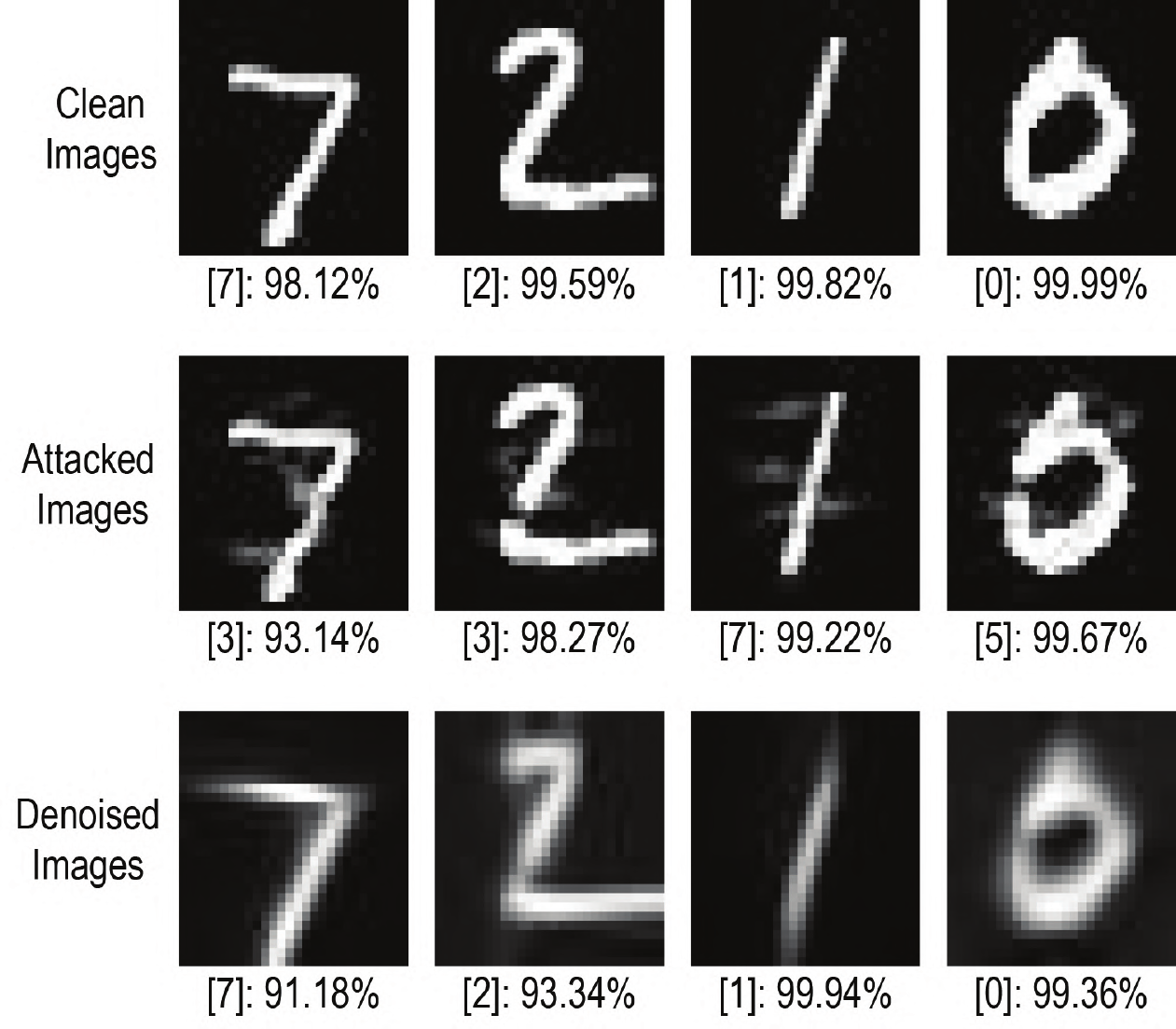}
    \caption{Visual illustration of MNIST images sorted  from top to bottom as clean images, attacked images using  \ac{bpda}-based \ac{pgd} under white-box attack at 1000 steps with \hbox{$\epsilon=0.3$}, and denoised images by the proposed defense method. The predicted class
label and its corresponding probability are provided for each image.}
    \label{fig:fig6}
\end{figure}
\begin{figure}[t!]
\centering
\includegraphics[scale=0.63]{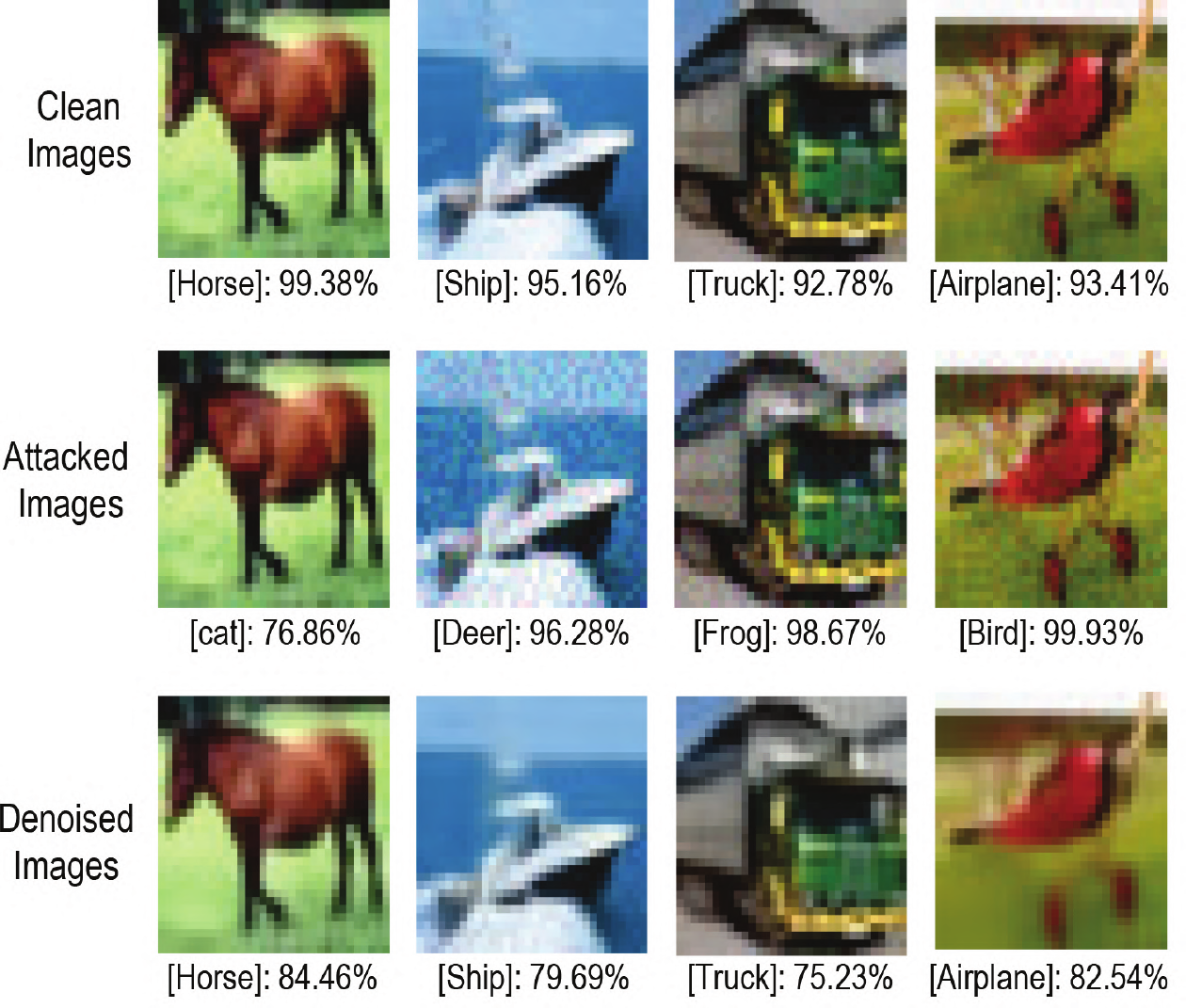}       
    \caption{Visual illustration of CIFAR-10 images sorted  from top to bottom as clean images, attacked images using \ac{bpda}-based \acs{pgd} under white-box attack at 1000 steps with \hbox{$\epsilon=0.3$}, and denoised by the proposed defense method. The predicted class
label and its corresponding probability are provided for each image.}
    \label{fig:fig7}
\end{figure}
\begin{figure*}[t!]
    \centering
\includegraphics[scale=0.65]{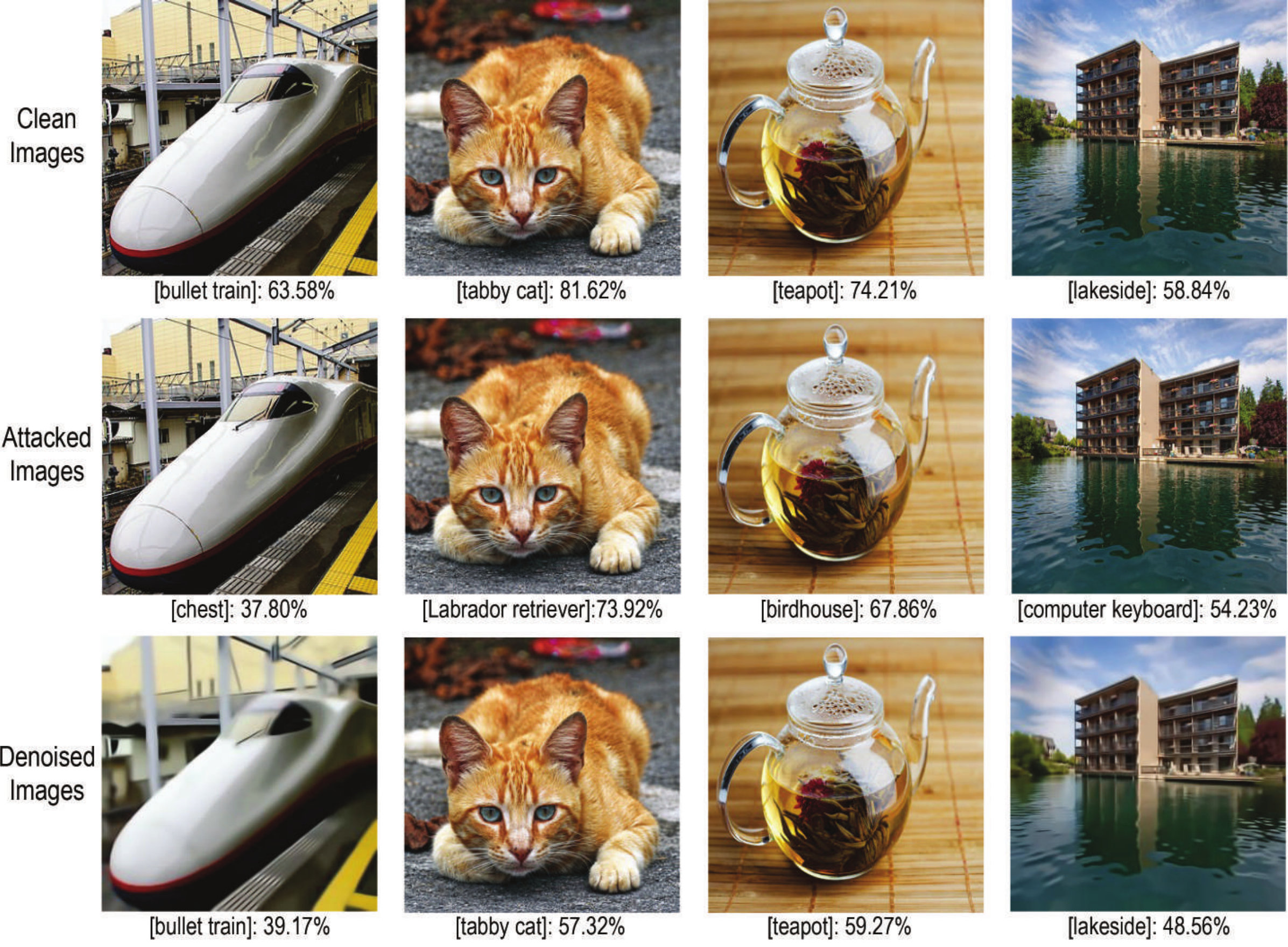}
    \caption{Visual illustration of Tiny-ImageNet images sorted  from top to bottom as  clean images, attacked images using  \ac{bpda}-based \ac{pgd} under white-box attack at 1000 steps with $\epsilon=0.03$, and denoised images by the proposed defense method. The predicted class
label and its corresponding probability are provided for each image.}
    \label{fig:fig8}
\end{figure*}

Figures \ref{fig:fig6}, \ref{fig:fig7} and \ref{fig:fig8} illustrate four images from  MNIST, CIFAR-10 and Tiny-ImageNet datasets, respectively. These images are illustrated in three configurations: clean, attacked with \ac{bpda}-based \ac{pgd} attack and denoised with the proposed denoiser block. We can notice that the adversarial perturbations are filtered by the denoiser block, while a slight blur is introduced to the denoised images. However, through the use of CNN-guided BM3D, the filtering is performed so that the introduced blur does not affect the classification performance.

\begin{figure*}[h!]
\centering
\subfigure[MNIST, without defense]{\label{fig:fig11a}\includegraphics[width=0.3\textwidth]{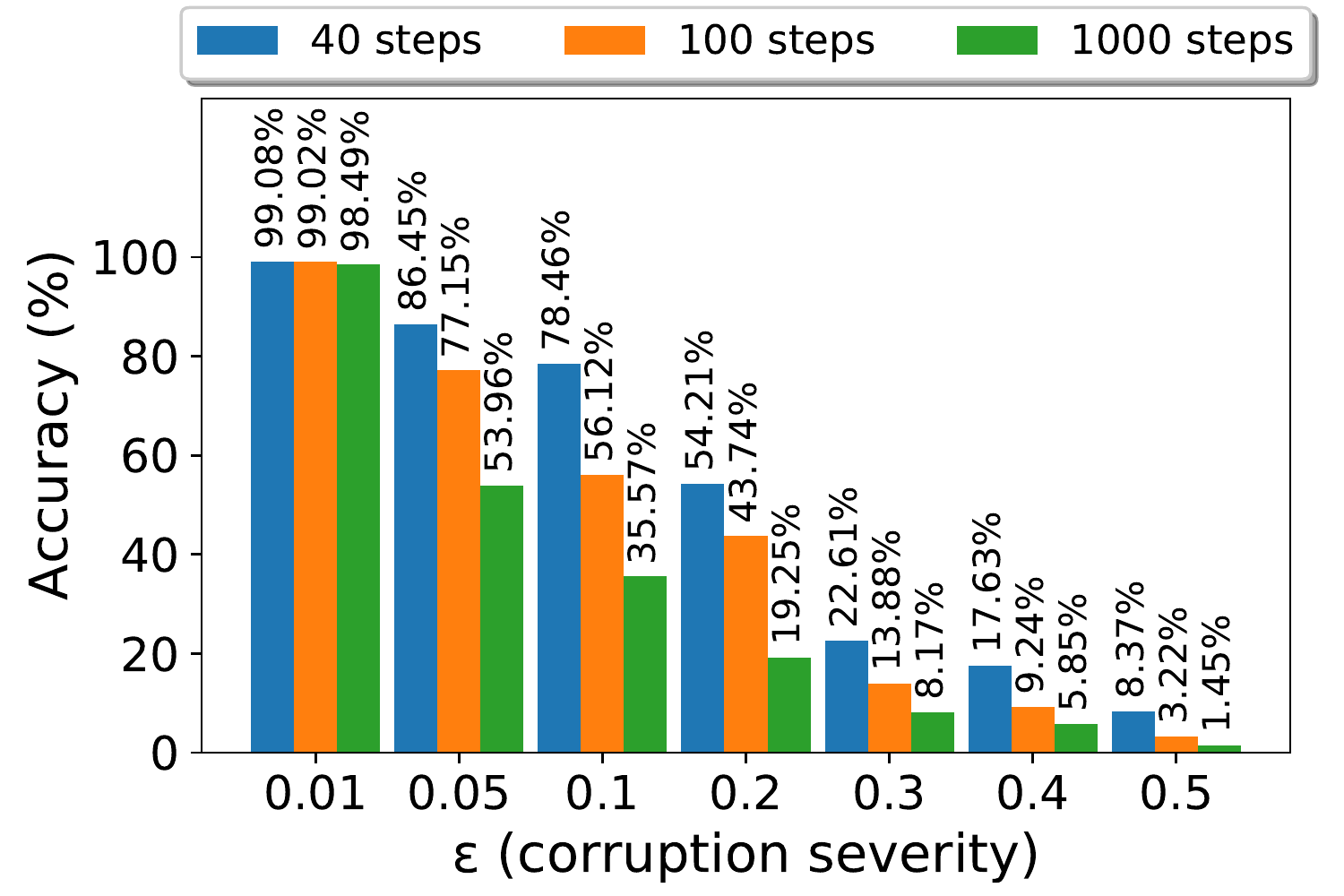}}
\subfigure[CIFAR-10, without defense]{\label{fig:fig12a}\includegraphics[width=0.3\textwidth]{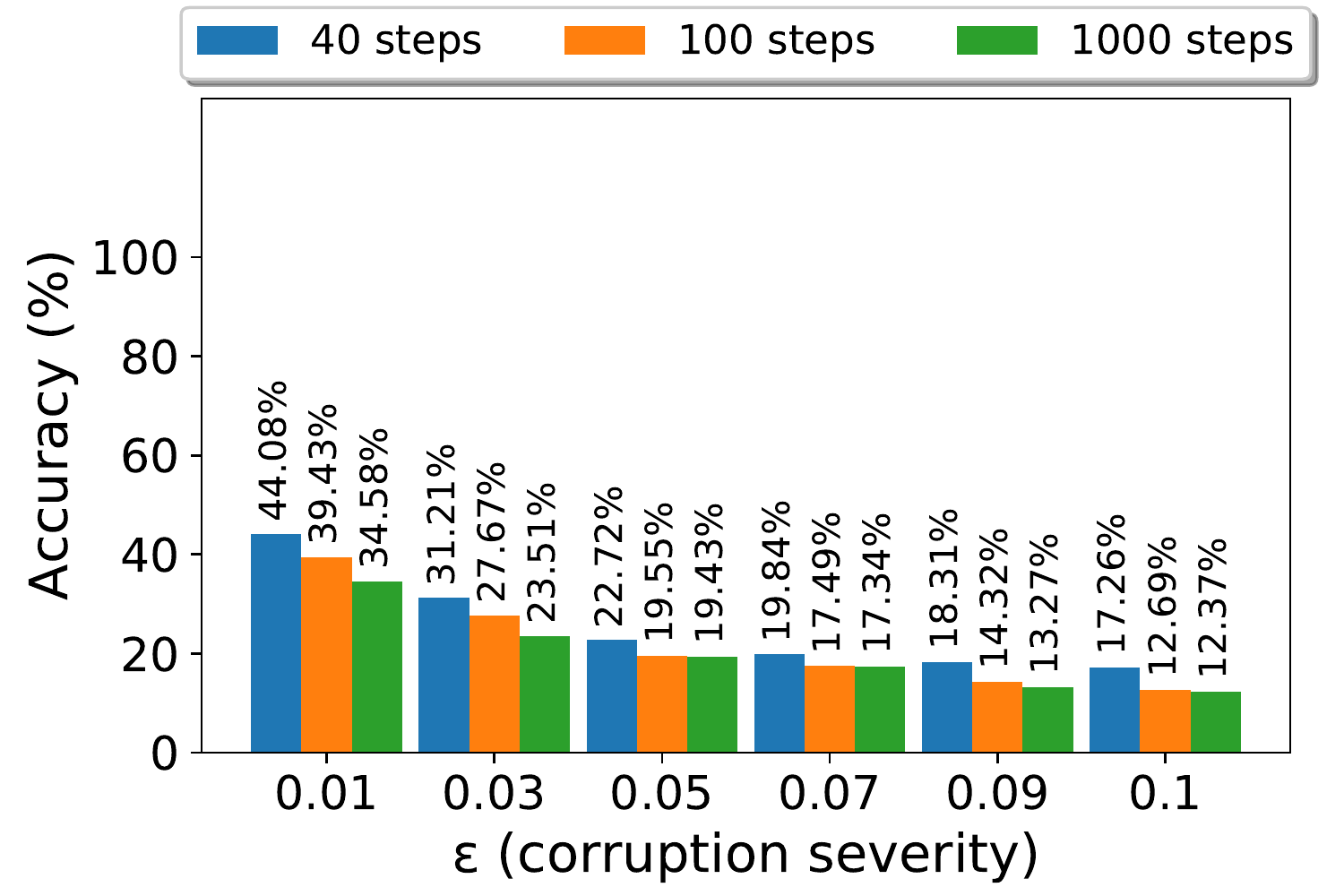}}
\subfigure[Tiny-ImageNet, without defense]{\label{fig:fig13a}\includegraphics[width=0.3\textwidth]{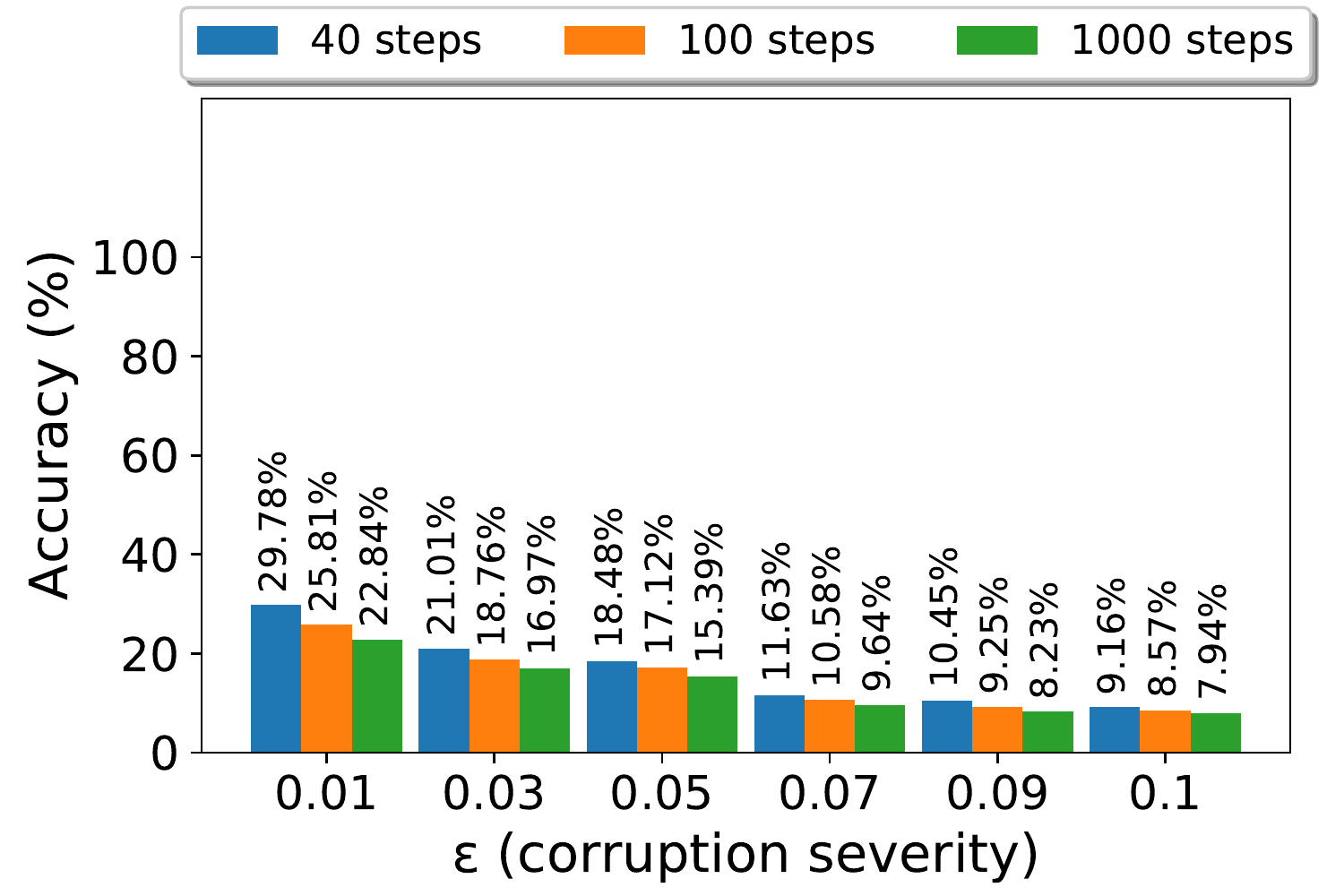}}

\subfigure[MNIST, with defense]{\label{fig:fig11b}\includegraphics[width=0.3\textwidth]{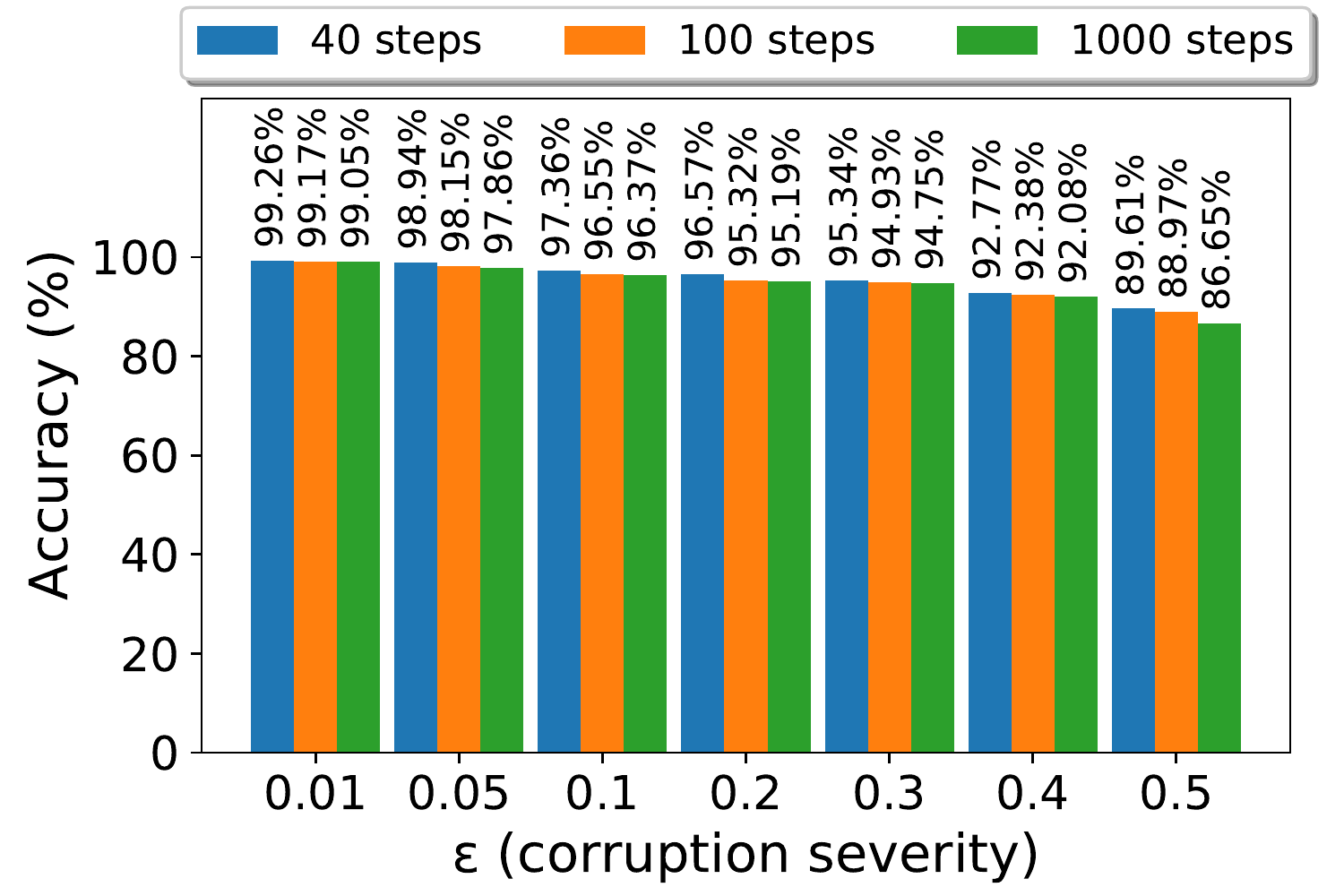}}
\subfigure[CIFAR-10, with defense]{\label{fig:fig12b}\includegraphics[width=0.3\textwidth]{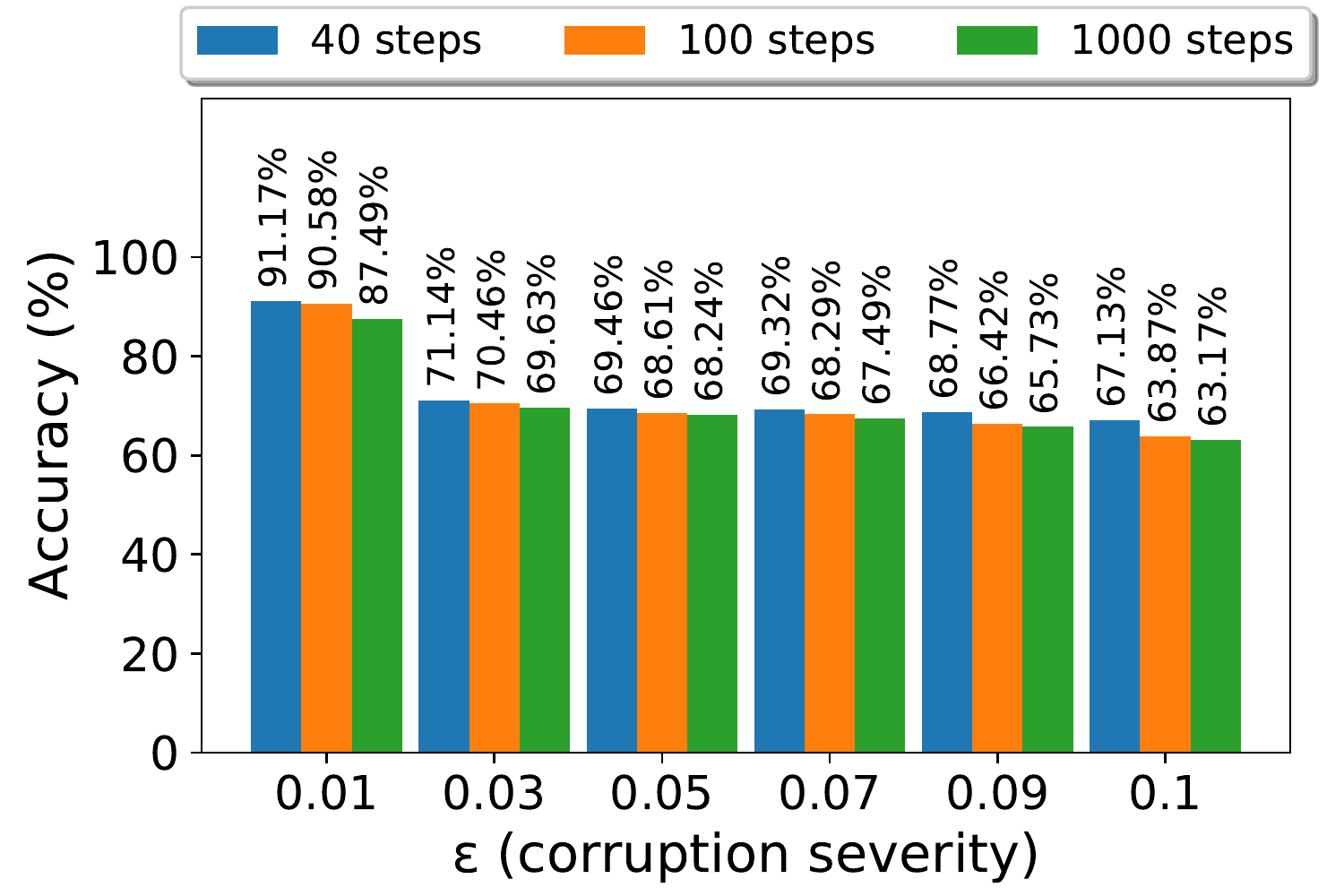}}
\subfigure[Tiny-ImageNet, with defense]{\label{fig:fig13b}\includegraphics[width=0.3\textwidth]{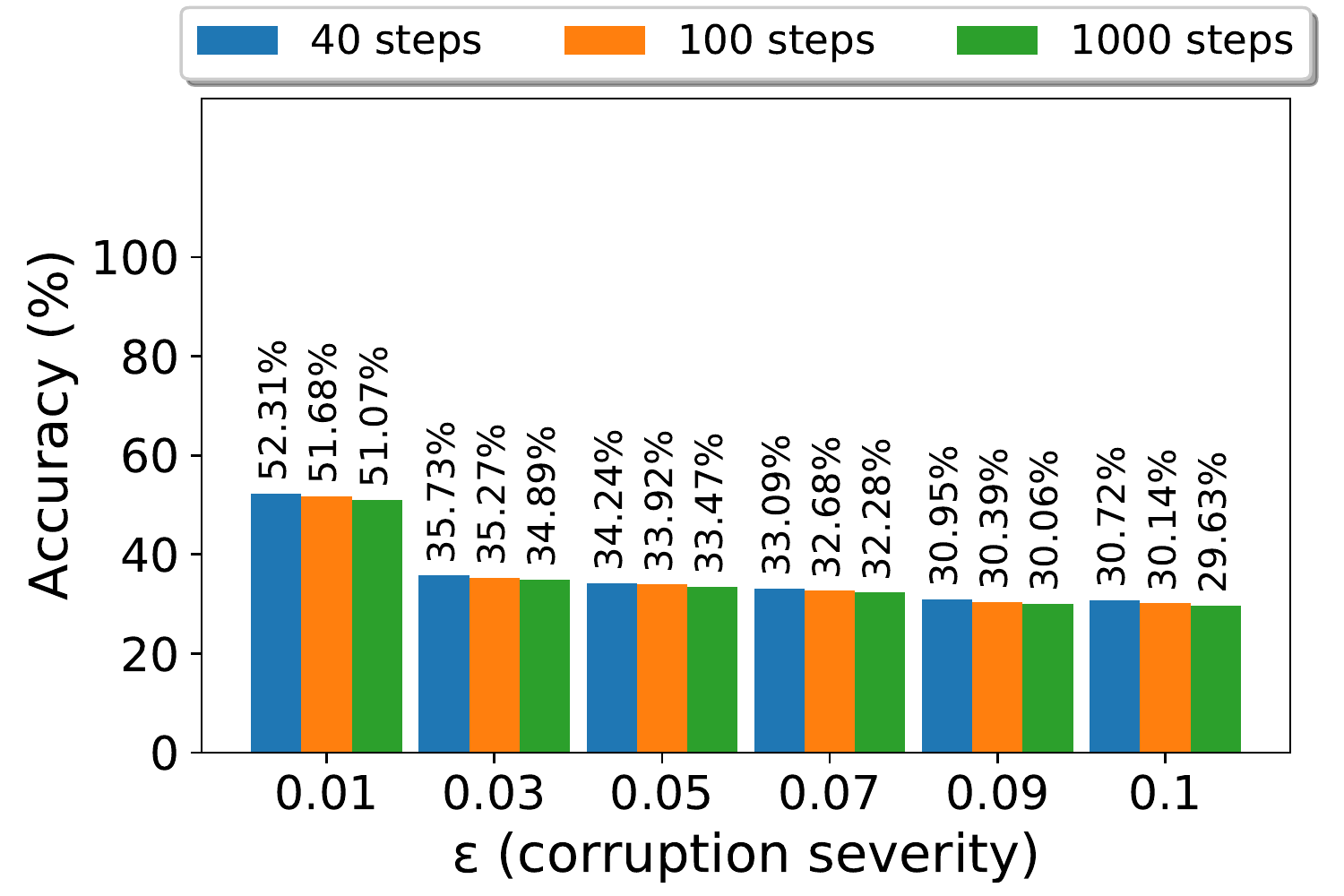}}

  \caption{Accuracy of the classifier against \ac{bpda}-based \ac{pgd} attack at three different steps for the three considered datasets.}
\label{fig:fig11}
\end{figure*}

Furthermore, in order to assess the performance of our defense technique in the worst case scenario, we changed the \ac{pgd} attack hyper-parameters under a white-box attack to see how the proposed method behaves against these perturbations. Figure~\ref{fig:fig11} illustrates the classification accuracy of the proposed technique against \ac{bpda}-based \ac{pgd} attack at different steps and perturbation magnitudes $\epsilon$. These figures clearly demonstrate the robustness of the proposed defense method  vis-à-vis the increasing in both the number of steps and the magnitude of perturbation. The accuracy of the proposed method slightly decreases when increasing the number of steps, while it remains robust regarding the magnitude increase. For MNIST dataset, the classifier performs very well even at high perturbation magnitudes, especially when $\epsilon>0.4$ which generates a very noisy image, making its classification difficult even by a human eye. Despite these hard conditions, our defense method achieves more than $86\%$ accuracy in the case of $\epsilon=0.5$ and $1000$ steps under white-box attack. For the color datasets, which are much more challenging to defend, our technique allows to increase the classification accuracy by approximately more than $40\%$ and $25\%$ for CIFAR-10 and Tiny-ImageNet, respectively.  



\begin{figure*}[h!]
\centering
\subfigure[\ac{fgsm} attack]{\label{fig:fig14a}\includegraphics[width=0.3\textwidth]{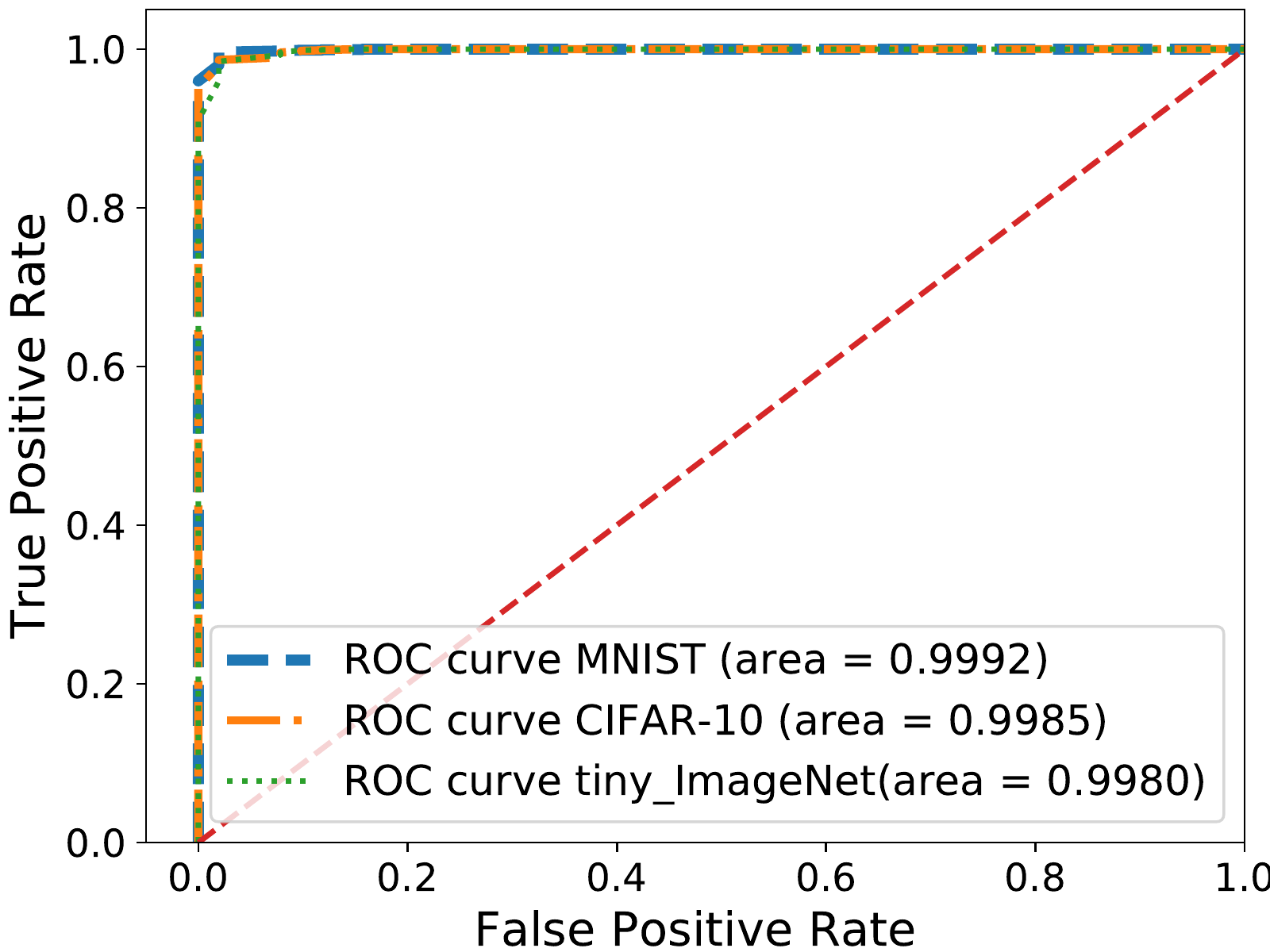}}
\subfigure[\acs{pgd} attack]{\label{fig:fig14b}\includegraphics[width=0.3\textwidth]{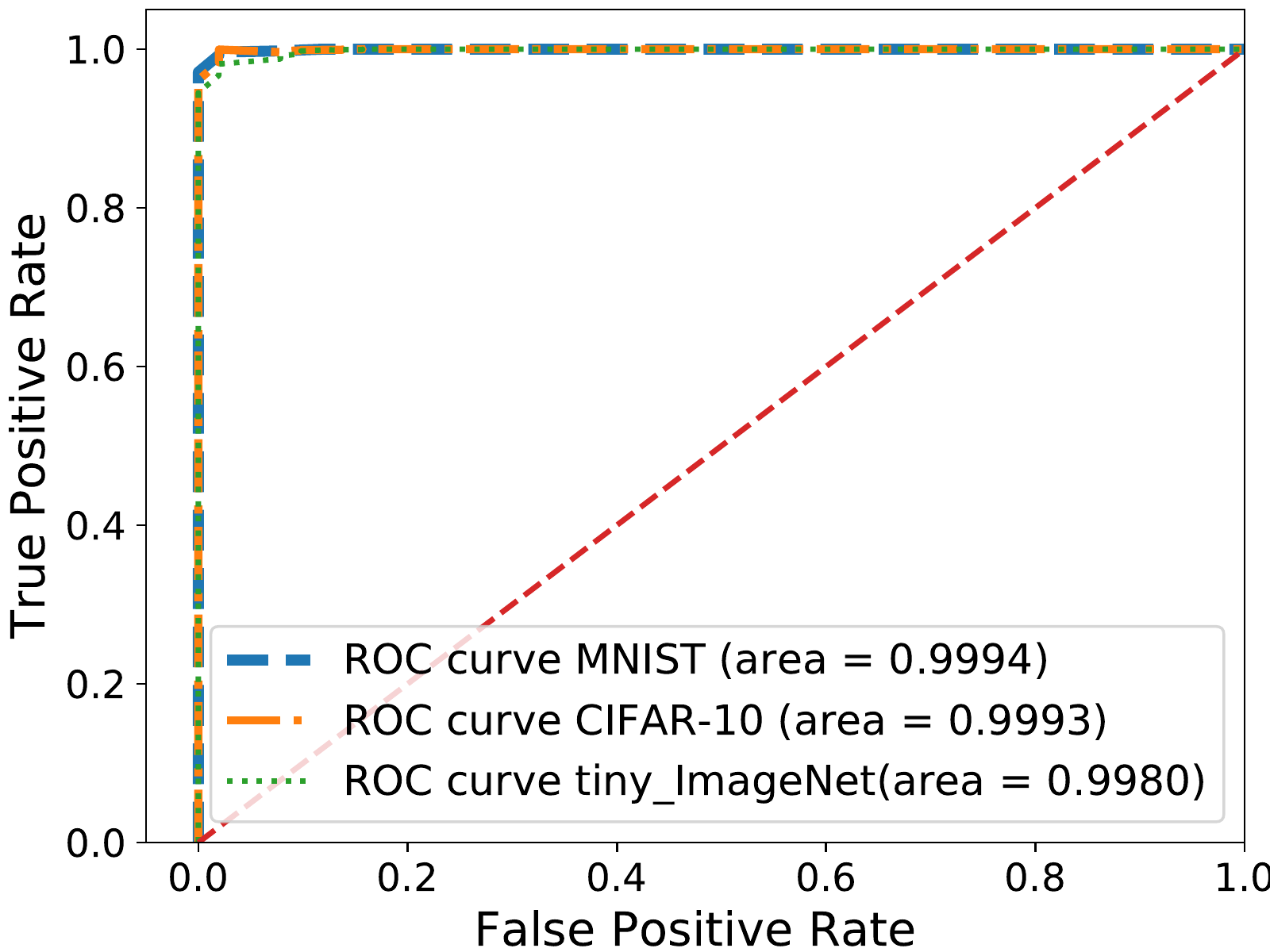}}
\subfigure[\acs{cw} attack]{\label{fig:fig14c}\includegraphics[width=0.3\textwidth]{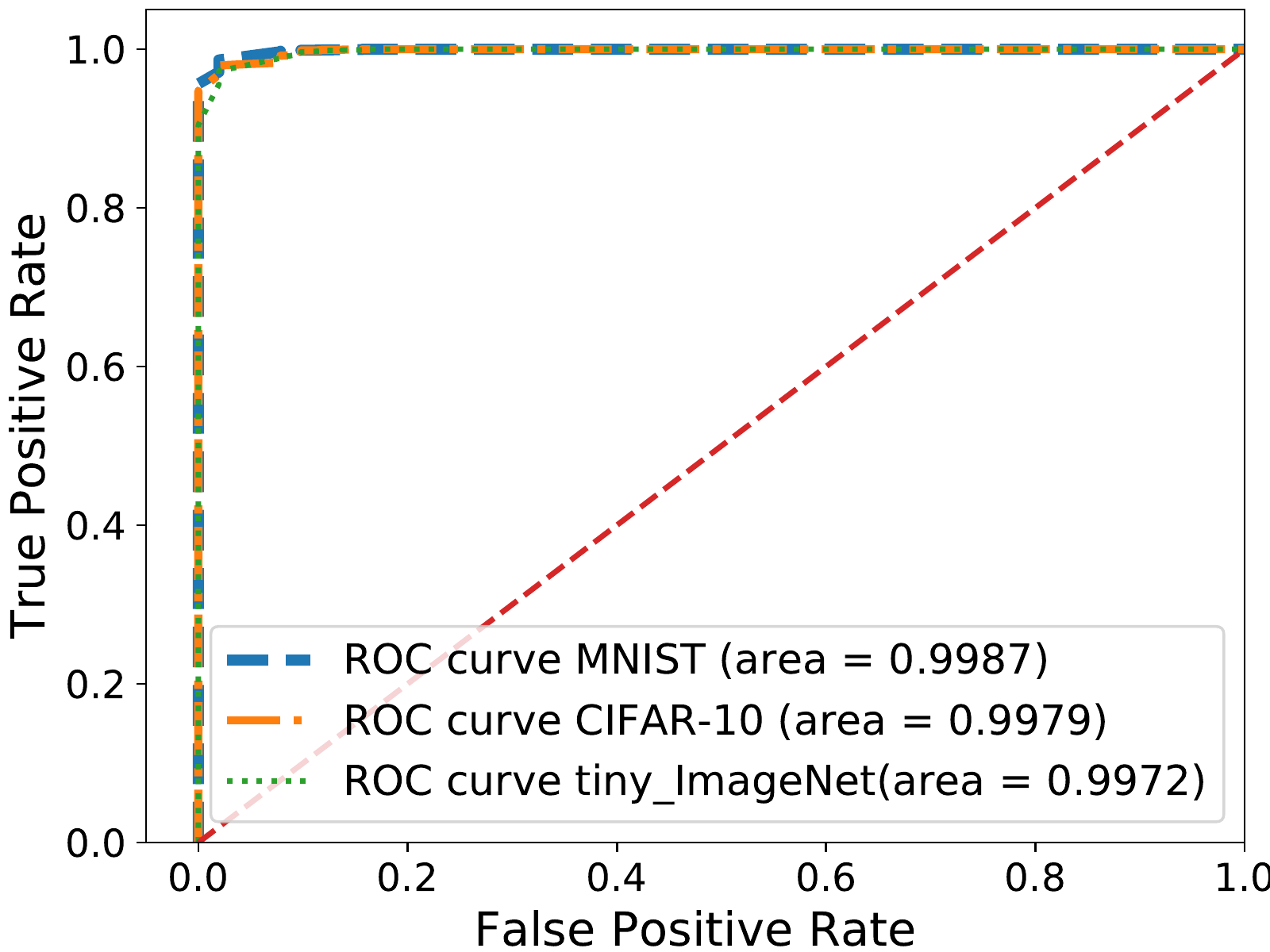}}
\caption{ROC performance of the detector block  against three attacks under white-box settings.}
\label{fig:fig14}
\end{figure*}

\subsubsection{Detector block performance}
The performance of the detector block is evaluated under the same conditions as the defense technique, while half of the images are benign and the other half are crafted using three white-box attacks including \ac{fgsm}, \ac{pgd} and \ac{cw}. Figure \ref{fig:fig14} summarizes the performance of our detector using \ac{roc} curves for different detection thresholds. The \ac{roc} curves are plotted with \ac{tpr} versus the \ac{fpr}. The area under the \ac{roc} curve (AUC) denotes the value measured by  the entire two-dimensional area below the entire \ac{roc} curve. The higher the AUC, the better the model is to predict clean images as clean and \acp{ae} as attacked. In the evaluation of the detector, we only selected images that were successfully attacked, i.e., $f_\theta(x^\prime) \neq f_\theta(x)$ , with their benign state. We reach an AUC of about over $0.9970$ for all datasets against the three used attacks, which is not far from an ideal detector.

Furthermore, Table \ref{tab:tab6} reports the detection accuracy against the three attacks under white-box settings, as well as the \ac{fp}, i.e, the detector classifies a benign input as an \ac{ae}. An efficient and robust detector must achieve high accuracy and at the same time a low \ac{fp}.  Our detector achieves $100\%$ detection accuracy for all white-box attacks and considered datasets, except for Tiny-ImageNet dataset against the strongest iterative \ac{cw} attack, where an accuracy of $98\%$ is reported. The later is an acceptable score because it is associated with a very low \ac{fp} of $2.3\%$.  


\begin{table}[t!]
\centering
\renewcommand{\arraystretch}{1.2} 
\caption{Accuracy and \acs{fp} of the detector against three attacks under white-box settings.\label{tab:tab6}}
\begin{tabular}{lllll}
\hline
 \multicolumn{1}{l}{\textbf{Dataset}}& \multicolumn{1}{c}{\textbf{\acs{fgsm}}}  &  \multicolumn{1}{c}{\textbf{\acs{pgd}}}  & \multicolumn{1}{c}{\textbf{\acs{cw}}}& \multicolumn{1}{c}{\textbf{\acs{fp}}(\%)}         \\ \hline
\multirow{1}{*}{\rotatebox[origin=c]{0}{MNIST}}           &\multicolumn{1}{c}{100\%}  
& \multicolumn{1}{c}{100\%}  
& \multicolumn{1}{c}{100\%} 
& \multicolumn{1}{c}{1.9\%}  \\  \hline 
\multirow{1}{*}{{CIFAR-10}}
&\multicolumn{1}{c}{100\%}  
& \multicolumn{1}{c}{100\%}
&\multicolumn{1}{c}{100\%}  
& \multicolumn{1}{c}{2.0\%}  \\   \hline 
\multirow{1}{*}{{Tiny-ImageNet}}
&\multicolumn{1}{c}{100\%}  
&\multicolumn{1}{c}{100\%}
&\multicolumn{1}{c}{98\%}
&\multicolumn{1}{c}{2.3\%} \\  \hline
\end{tabular}
\end{table}
\section{Conclusion}
\label{sec:conclusion}
\label{sec5}
In this paper, we have proposed a novel two-stage framework to defend against \acp{ae} involving detection of \acp{ae} followed by denoising. The detector relies on \ac{nss} of the input image, which are altered by the presence of adversarial perturbations. The samples detected as malicious are then processed by the denoiser with an adaptive \ac{bm3d} filter to project them back into there original manifold. The parameters of the \ac{bm3d} filter used to process the attacked image are estimated by a \ac{cnn}.         

The performance of the proposed defense method is extensively evaluated against black-box, grey-box and white-box attacks on three standard datasets. The experimental results showed the effectiveness of our defense method in improving the robustness of \acp{dnn} in the presence of \acp{ae}. Moreover, the efficiency of the detector with high detection accuracy and low \ac{fp} helps to preserve the classification accuracy of the \ac{dnn} model on clean images. \addcomment{As future work, we plan to extend the proposed defense method to deal with adversarial attacks in the physical world.}



%
\section*{Conflict of interest}
The authors declare that they have no conflict of interest.
\vspace{-5mm}



\end{document}